\documentclass[10pt]{article}

\usepackage{graphicx}
\usepackage{amsmath}
\usepackage{amssymb}
\usepackage{caption}
\usepackage{subcaption}
\usepackage{algorithm}
\usepackage{algpseudocode}
\usepackage[acronym]{glossaries}

\graphicspath{{figures/}}

\usepackage[pagebackref=false,breaklinks=true,letterpaper=true,bookmarks=false]{hyperref}

\newacronym{psnr}{PSNR}{peak signal-to-noise ratio}
\newacronym{he}{HE}{histogram equalization}
\newacronym{bbhe}{BBHE}{brightness preserving bi-histogram equalization}
\newacronym{dsihe}{DSIHE}{dualistic sub-image histogram equalization}
\newacronym{clahe}{CLAHE}{contrast-limited adaptive histogram equalization}
\newacronym{octm}{OCTM}{optimal contrast-tone mapping}
\newacronym{hvs}{HVS}{human visual system}
\newacronym{hdr}{HDR}{high dynamic range}
\newacronym{dct}{DCT}{discrete cosine transform}
\newacronym{eme}{EME}{enhancement measure}
\newacronym{larcotm}{LARCOTM}{locally adaptive rank-constrained optimal tone mapping}
\newacronym{iid}{i.i.d.}{independent and identically distributed}
\newacronym{1d}{1D}{one dimensional}
\newacronym{2d}{2D}{two dimensional}
\newacronym{qf}{QF}{quality factor}
\newacronym{nns}{NNS}{nonlocal self-similarity}
\newacronym{svt}{SVT}{singular value thresholding}
\newacronym{mse}{MSE}{mean square error}
\newacronym{rmse}{RMSE}{root mean square error}
\newacronym{nqcs}{NQCS}{narrow quantization constraint set}

\makeglossaries

\newlength{\subfigwidth}

\newcommand{\D}[1]{\ensuremath{\operatorname{d}\!{#1}}}

\begin{document}

\title{Quality Adaptive Low-Rank Based JPEG Decoding with
  Applications}

\author{Xiao Shu \qquad Xiaolin Wu \\
  McMaster University
}

\date{}

\maketitle

\begin{abstract}
  Small compression noises, despite being transparent to human eyes,
  can adversely affect the results of many image restoration
  processes, if left unaccounted for.  Especially, compression noises
  are highly detrimental to inverse operators of high-boosting
  (sharpening) nature, such as deblurring and superresolution against
  a convolution kernel.  By incorporating the non-linear DCT
  quantization mechanism into the formulation for image restoration,
  we propose a new sparsity-based convex programming approach for
  joint compression noise removal and image restoration.  Experimental
  results demonstrate significant performance gains of the new
  approach over existing image restoration methods.
\end{abstract}

\section{Introduction}

Image restoration is to improve the quality of acquired image data in
preparation for higher level vision tasks.  It remains a very active
research area because the precision and success of many computer
vision algorithms, such as registration, recognition, detection,
classification, matting, retrieval, etc., depend on the quality of the
input image.  But much to our surprise, in the very large existing
body of research literature on image restoration, very little study
has been reported on how compression noises affect the performances of
various image restoration processes, such as deconvolution,
superresolution, etc.  All published works of image restoration,
except few papers explicitly on the topic of combating compression
artifacts (a.k.a., soft decoding), assumed the input image data to be
uncompressed or mathematically losslessly compressed.  This long-time
tradition is, unfortunately, an operational convenience in contrary to
the real world settings.  In most practical scenarios, particularly
those of consumer applications, the input images are compressed in
\gls{dct} domain with some loss of fidelity.  For practical systems
constrained by bandwidth and storage economy, lossy compression is
inevitable because mathematically invertible image coding typically
achieves only roughly 2:1 compression ratio,
still leaving the image file size too large to handle.

Granted, after years of research, development and investment,
international compression standards such as JPEG, H.264, HEVC, etc.,
can offer very high reconstruction quality to the level of perceptual
transparency; namely, naked eyes cannot discern any difference between
the original and the decompressed images.  But as demonstrated by this
work, small compression noises, despite being transparent to human
eyes, can adversely affect the results of many image restoration
processes, if left unaccounted for.  Especially, compression noises
are highly detrimental to inverse operators of high-boosting
(sharpening) nature, such as deblurring and superresolution against a
convolution kernel.

\begin{figure}
  \centering

  \setlength{\subfigwidth}{\linewidth}

  \begin{subfigure}[b]{\subfigwidth}
    \centering
    \includegraphics[width=0.32 \textwidth]{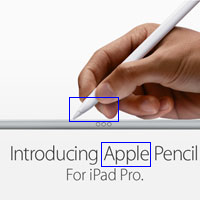}
    \includegraphics[width=0.32 \textwidth]{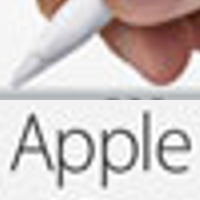}
    \includegraphics[width=0.32 \textwidth]{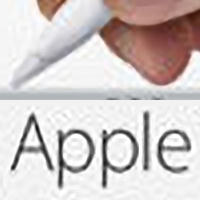}
    \caption{Original image}
    \label{fig:pencil_orig}
  \end{subfigure}

  \begin{subfigure}[b]{\subfigwidth}
    \centering
    \includegraphics[width=0.32 \textwidth]{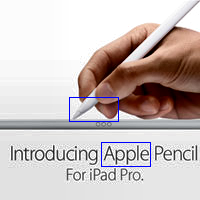}
    \includegraphics[width=0.32 \textwidth]{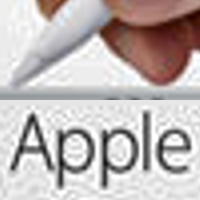}
    \includegraphics[width=0.32 \textwidth]{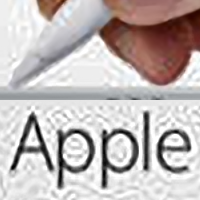}
    \caption{Enhanced by CLAHE \cite{pizer1987adaptive}}
    \label{fig:pencil_clahe}
  \end{subfigure}

  \begin{subfigure}[b]{\subfigwidth}
    \centering
    \includegraphics[width=0.32 \textwidth]{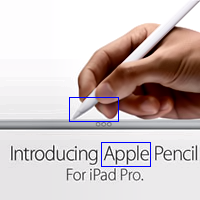}
    \includegraphics[width=0.32 \textwidth]{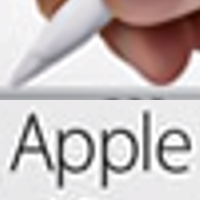}
    \includegraphics[width=0.32 \textwidth]{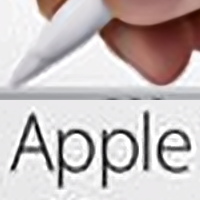}
    \caption{Enhanced by the proposed technique}
    \label{fig:pencil_ours}
  \end{subfigure}

  \caption{JPEG compression artifacts become highly objectionable
    after image enhancement or image size magnification, or both.  The
    second and third sub-figures from left are regions up-scaled by
    bi-cubic interpolation and A+ \cite{timofte2014a+}, respectively.}

  \label{fig:pencil}
\end{figure}

The omission of compression noises in the design of image restoration
algorithms is seemingly due to the fact that the compression noises
are much more difficult to model than other degradation sources, e.g.,
motion blur and sensor noises.  The compression-induced quality
degradation is more pronounced for compound document images, which are
characterized by the embedding of graphics arts or texts into an
acquired photograph, as exemplified by Figure~\ref{fig:pencil}.  The
non-linearity of quantization operations in image compression systems
makes quantization noises image dependent, far from being white and
independent, as commonly assumed by most researchers in the field of
image restoration.

The contributions of this paper are two folds. First, we analyze the
nature of quantization noises in the DCT domain, in which most popular
JPEG and H.264/HEVC compression standards operate.  In particular, we
find that the quantization errors of DCT coefficients exhibit complex
behaviours after being mapped back into the spatial domain. These
behaviours are highly sensitive to quantization precision,
the amplitude and phase of the input image signal.  Second, we manage
to incorporate the non-linear DCT quantization mechanism into the
inverse problem formulation for image restoration.  Specifically, we
propose a new sparsity-based convex programming approach for joint
quantization noise removal and restoration, verify the efficacy of the
proposed approach for the tasks of deblurring and super-resolving
DCT-domain compressed images, and demonstrate significant performance
gains of the new approach over existing deblurring and superresolution
methods.

\section{Quantization Error in DCT Domain}
\label{sec:dct}

The process of capturing, storing and displaying a digital image is
far from perfect; it often introduces objectionable errors, such as
motion blur, lens distortion, moir\'{e} pattern, sensor noise,
compression noise, etc., into the final reproduction of a scene.  Some
errors are independent to and statistically distinct from signal.  For
example, sensor noise can be modelled as random variables following an
\gls{iid} Gaussian distribution, while true signal has repetitive
patterns hence sparse in some basis \cite{mairal2009non,
  dong2011sparsity}.  By exploiting this statistical difference
between signal and sensor noise, denoising techniques can effectively
separate signal and noise in a given noisy observation
\cite{gu2014weighted}.  Compression noises, on the other hand, are
much more difficult to model than other degradation sources, e.g.,
motion blur and sensor noises. The non-linearity of quantization
operations in image compression systems makes quantization noises
image dependent, far from being white and independent.

\begin{figure}
  \centering

  \setlength{\subfigwidth}{0.24\linewidth}

  \begin{subfigure}[b]{\subfigwidth}
    \includegraphics[width=\textwidth]{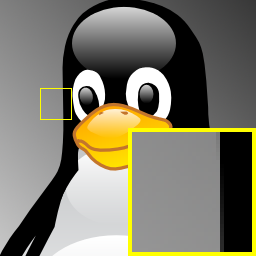}
    \caption{\footnotesize Original}
    \label{fig:jpeg_orig}
  \end{subfigure}
  \begin{subfigure}[b]{\subfigwidth}
    \includegraphics[width=\textwidth]{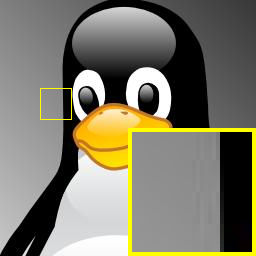}
    \caption{\footnotesize JPEG ($\mbox{QF}=75$)}
    \label{fig:jpeg_75}
  \end{subfigure}
  \begin{subfigure}[b]{\subfigwidth}
    \includegraphics[width=\textwidth]{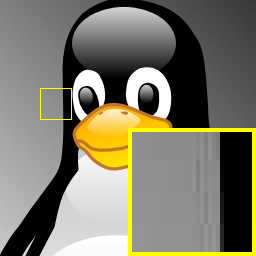}
    \caption{\footnotesize JPEG ($\mbox{QF}=50$)}
    \label{fig:jpeg_50}
  \end{subfigure}
  \begin{subfigure}[b]{\subfigwidth}
    \includegraphics[width=\textwidth]{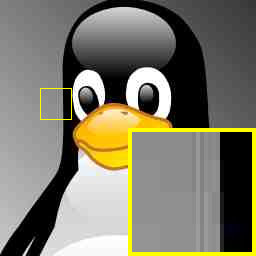}
    \caption{\footnotesize JPEG ($\mbox{QF}=25$)}
    \label{fig:jpeg_25}
  \end{subfigure}

  \caption{JPEG compression noise is corrected with signal.}

  \label{fig:jpeg}
\end{figure}

In main stream DCT-based compression systems, the encoding of signal
$\boldsymbol{x}$ is a three-step process.  1.~The discrete cosine
transform $T$ is performed on signal $\boldsymbol{x}$; 2.~the
transformed signal $T(\boldsymbol{x})$ is subject to quantization $Q$;
3.~the quantized version $(Q \circ T)(\boldsymbol{x})$ is coded by an
entropy coder $C$, resulting the code stream $(C \circ Q \circ
T)(\boldsymbol{x})$ for storage or transmission.  The decoding process
reverses the above three-step encoding process and generates the
decompressed signal
\begin{equation}
  \boldsymbol{\hat{x}} = (T^{-1} \circ Q^{-1} \circ C^{-1})
  ((C \circ Q \circ T)(\boldsymbol{x})).
\end{equation}
In this closed loop, the entropy decoder $C^{-1}$ and the inverse
transform $T^{-1}$ are invertible operators, namely, $C^{-1} \circ
C=I$, $T^{-1} \circ T = I$, but the dequantization operator $Q^{-1}$
is not.  The approximation error due to $Q^{-1} \circ Q \neq I$ is
aggravated and complicated by the non-linearity of the quantization
operation $Q$.  In the interest of gaining compression performance,
the quantizer $Q$ inclines to demote or outright discard
high-frequency DCT coefficients.  Setting high frequency components of
$\boldsymbol{x}$ to zero causes periodic ringing artifacts in the
reconstructed signal $\boldsymbol{\hat{x}}$, which are easy to
perceive as demonstrated in Figure~\ref{fig:jpeg}.  In this set of
JPEG-decompressed images, the ringing artifacts not only accompany
sharp edges in close proximity and they also agree with the image
signal in orientation; in other words, the quantization noises are
correlated with the image signal.

Unlike other noise mechanisms in image or video restoration,
compression noises are not random in the sense that coding blocks of
similar high-frequency contents tend to have similar ringing
artifacts.  As a result, a particular artifact pattern may occur
repetitively in a pixel vicinity.  Such signal-dependent noises may
resist the treatment of sparsity-based denoising techniques, because
the assumption that only the signal as self-similarity is no longer
valid.



\begin{figure}
  \centering

  \setlength{\subfigwidth}{0.22\linewidth}

  \begin{subfigure}[b]{\linewidth}
    \centering
    \includegraphics[width=1.1364\subfigwidth]{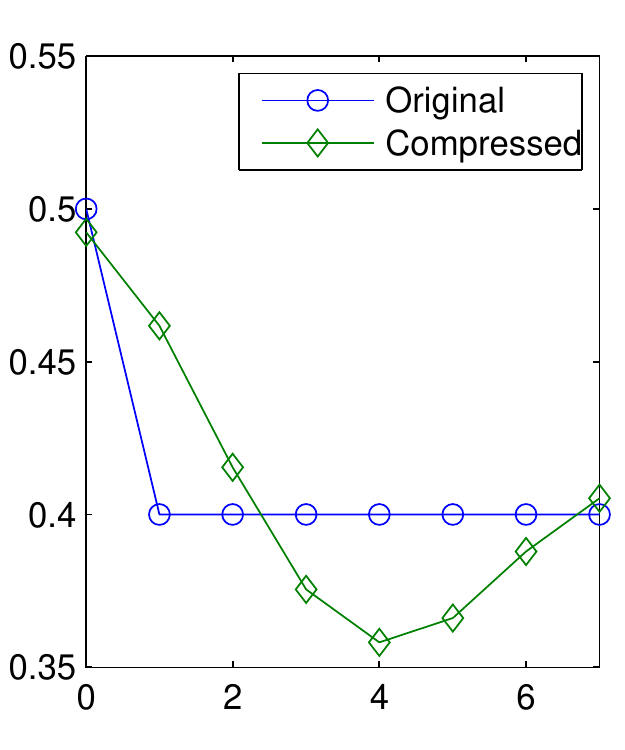}
    \includegraphics[width=\subfigwidth]{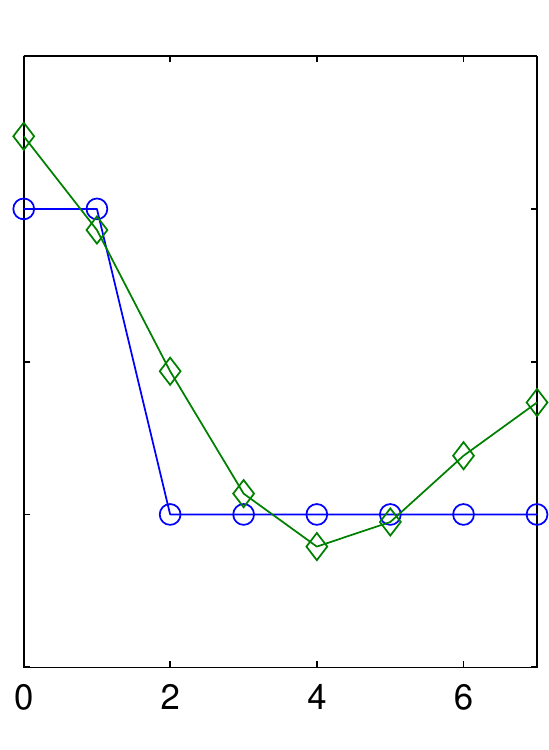}
    \includegraphics[width=\subfigwidth]{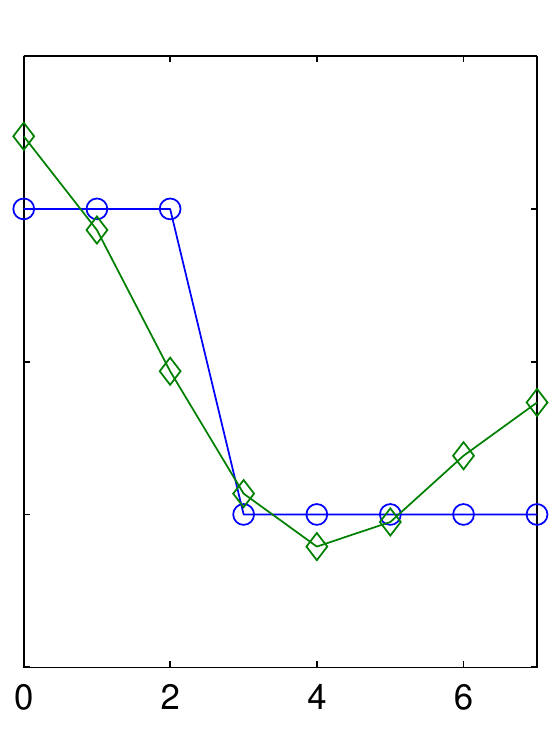}
    \includegraphics[width=\subfigwidth]{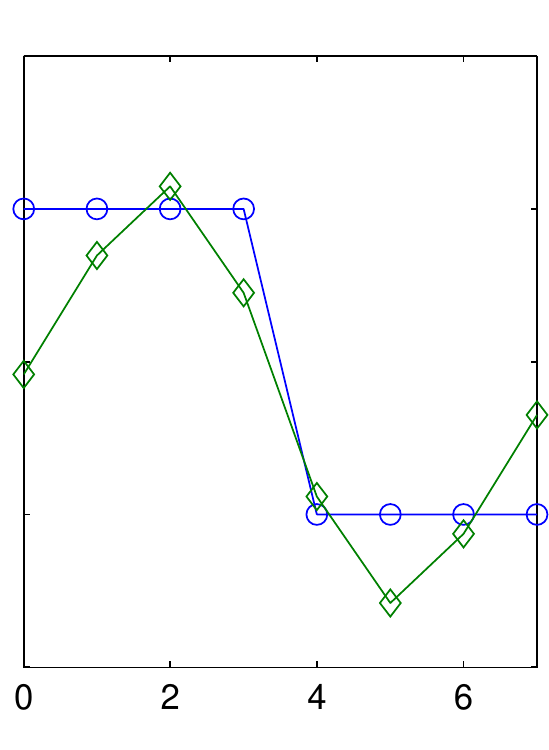}
    \caption{$\mbox{Amplitude}=0.1$, $\mbox{QF}=45$.}
  \end{subfigure}

  \begin{subfigure}[b]{\linewidth}
    \centering
    \includegraphics[width=1.1364\subfigwidth]{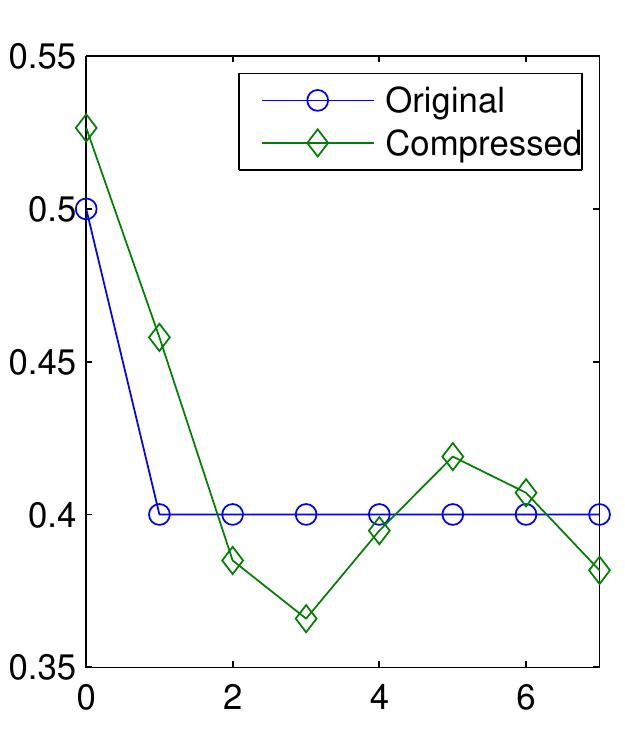}
    \includegraphics[width=\subfigwidth]{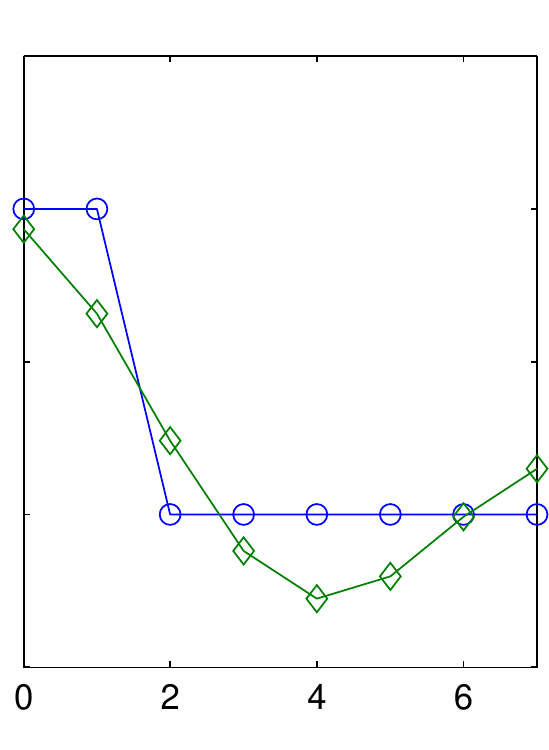}
    \includegraphics[width=\subfigwidth]{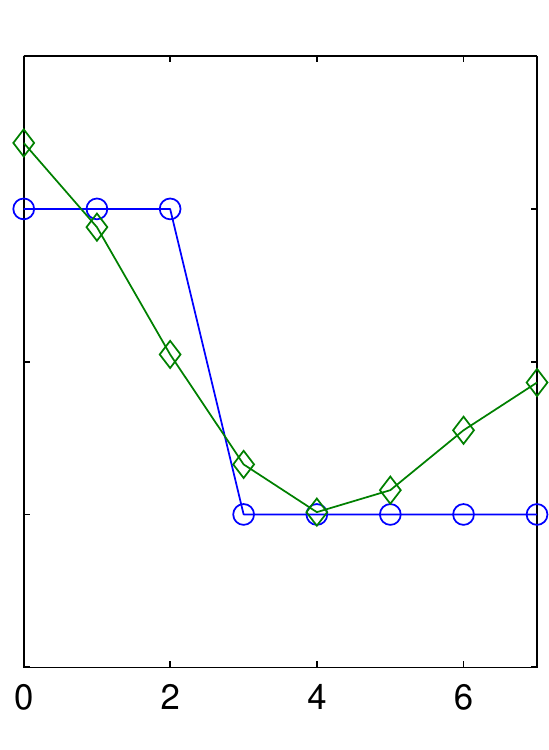}
    \includegraphics[width=\subfigwidth]{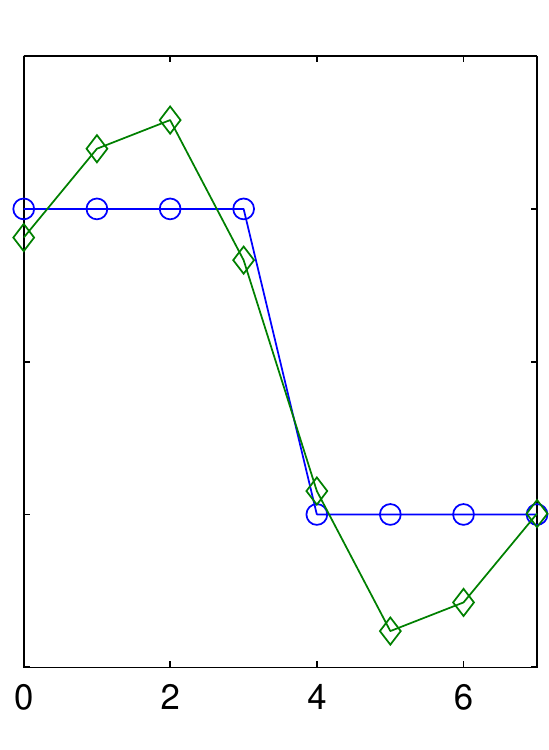}
    \caption{$\mbox{Amplitude}=0.1$, $\mbox{QF}=50$.}
  \end{subfigure}

  \begin{subfigure}[b]{\linewidth}
    \centering
    \includegraphics[width=1.1364\subfigwidth]{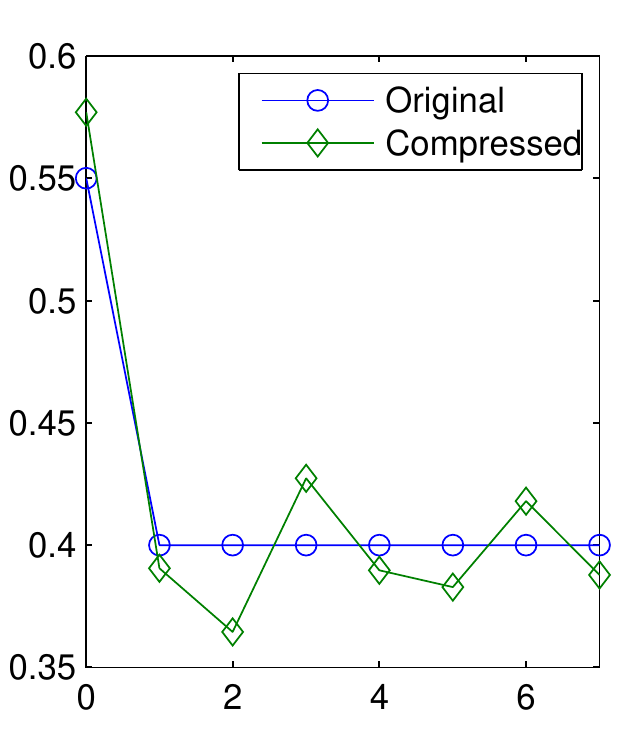}
    \includegraphics[width=\subfigwidth]{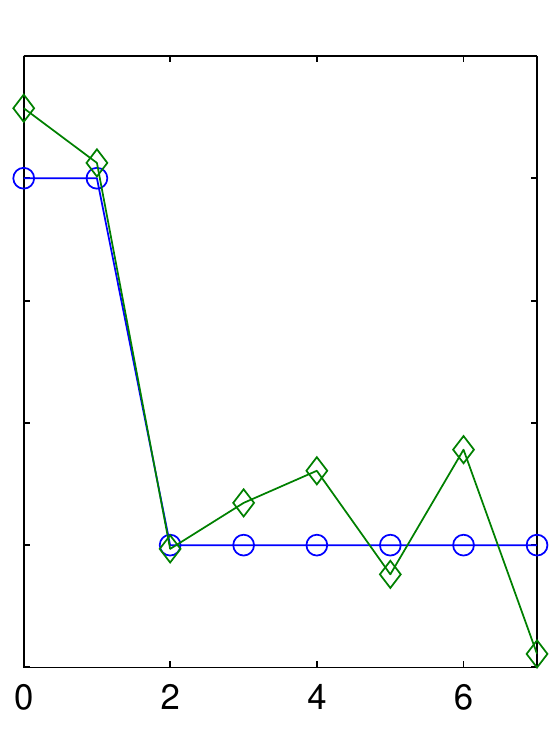}
    \includegraphics[width=\subfigwidth]{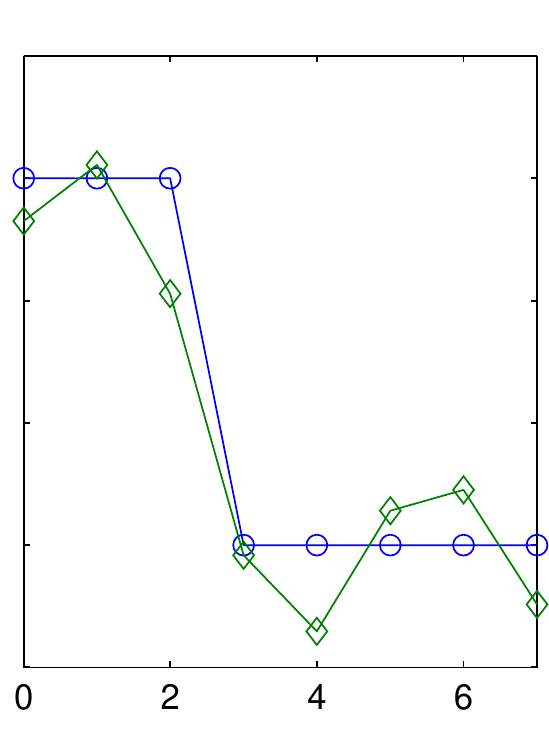}
    \includegraphics[width=\subfigwidth]{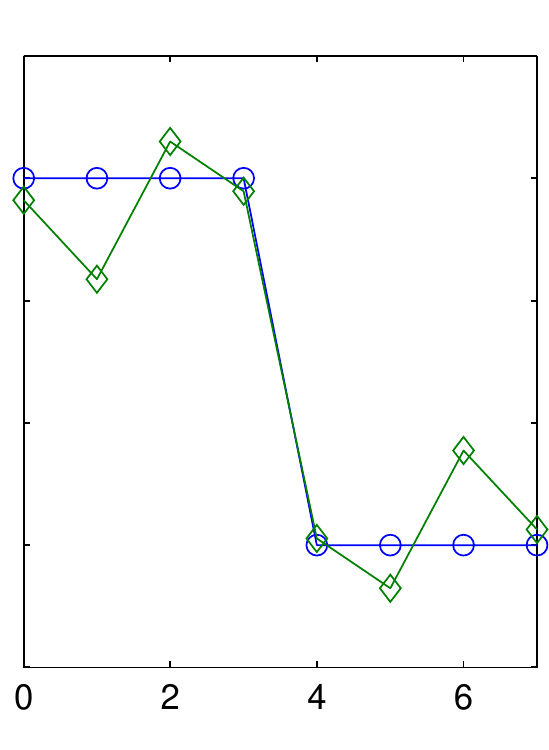}
    \caption{$\mbox{Amplitude}=0.15$, $\mbox{QF}=50$.}
  \end{subfigure}

  \caption{\gls{dct} quantization noise appears drastically different
    with small changes in QF, signal phase or amplitude.}

  \label{fig:noise}
\end{figure}

\begin{figure}
  \centering

  \setlength{\subfigwidth}{0.24\linewidth}

  \begin{subfigure}[b]{\subfigwidth}
    \includegraphics[width=\textwidth]{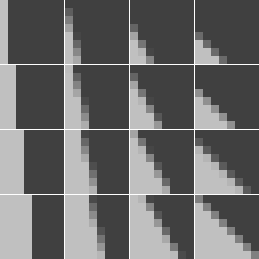}
    \caption{\footnotesize Original}
    \label{fig:jpeg_orig}
  \end{subfigure}
  \begin{subfigure}[b]{\subfigwidth}
    \includegraphics[width=\textwidth]{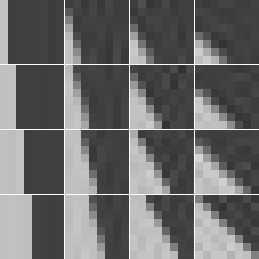}
    \caption{\footnotesize JPEG ($\mbox{QF}=75$)}
    \label{fig:jpeg_75}
  \end{subfigure}
  \begin{subfigure}[b]{\subfigwidth}
    \includegraphics[width=\textwidth]{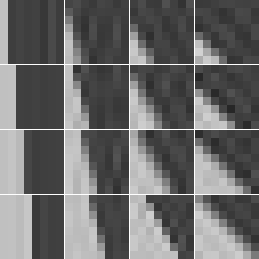}
    \caption{\footnotesize JPEG ($\mbox{QF}=50$)}
    \label{fig:jpeg_50}
  \end{subfigure}
  \begin{subfigure}[b]{\subfigwidth}
    \includegraphics[width=\textwidth]{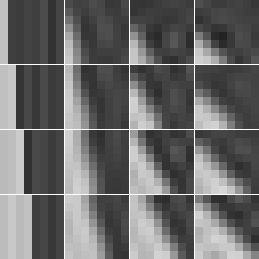}
    \caption{\footnotesize JPEG ($\mbox{QF}=25$)}
    \label{fig:jpeg_25}
  \end{subfigure}

  \caption{Besides QF, JPEG compression noise is sensitive to the
    angle and phase of the signal.}

  \label{fig:angle}
\end{figure}

There is yet another complication in modeling and removing DCT
quantization errors.  That is, the same signal structure can, after
through the loop of compression $C \circ Q \circ T$ and decompression
$T^{-1} \circ Q^{-1} \circ C^{-1}$, exhibit much varied temporal or
spatial patterns, with even immaterial changes in the phase or
amplitude of the input signal, and in compression \gls{qf}.  This high
sensitivity and nonlinearity of error patterns are depicted
graphically in Figure~\ref{fig:noise}.  In this example, the
reconstructed versions of the same \gls{1d} unit pulse signal of minor
linear-type alterations, such as shifting and scaling, behave
drastically differently.  In \gls{2d} image, there are more factors
affecting quantization noise.  Figure~\ref{fig:angle} shows the
quantization effects on image blocks of the same sharp edge but
different phases and angles.  In each case, there are visible false
lines parallel to the edge, however, the position, strength and sign
of the compression noise vary with a small change in angel or phase.
Since there are so many factors affecting the quantization error,
using learning based techniques to build a map from noisy observation
to true signal for each scenario is impractical.


\section{DCT Quantization Error Model}
\label{sec:model}

\Gls{dct} based lossy compression techniques realize data volume
reduction by trading off the accuracy of the DCT-domain representation
of the input signal through quantization.  By the definition of
\gls{dct}, the $k$-th \gls{dct} coefficient of \gls{1d} signal $x_0,
\ldots, x_{N-1}$ is,
\begin{equation}
  X_k = \sum_{n=0}^{N-1} x_n \cos \left[ \frac{\pi}{N}
    \left( n + \frac{1}{2} \right) k \right].
  \label{eq:dctdef}
\end{equation}
After quantization, the true value of $X_k$ is commonly estimated as,
\begin{equation}
  \hat{X}_k = \lfloor X_k / q_k + 0.5 \rfloor \cdot q_k,
\end{equation}
where $q_k$ is quantization interval for the $k$-th DCT coefficient.
In general, $q_k$ is set to decrease with $k$, due to the fact that
most energy of a signal is commonly concentrated in low frequency
components.

Using Fourier transform, DCT can be approximated in continuous domain
as follows,
\begin{align}
  X_k
  &= N \sum_{n=0}^{N-1} \frac{1}{N} x_n \cos
    \left( 2\pi \cdot \frac{n + \frac{1}{2}}{N} \cdot \frac{k}{2} \right)
    \nonumber
  \\
  &\approx N \int_0^1 f(t) \cos \left(2 \pi t \frac{k}{2} \right) \D{t}
    \nonumber
  \\
  &= \operatorname{Re} \left[ \frac{N}{2} \int_{-\infty}^\infty f(t)
    e^{-2 \pi t \frac{k}{2}} \D{t} \right]
    \nonumber
  \\
  &= \operatorname{Re} \left[ \frac{N}{2}
    F \left( \frac{k}{2} \right) \right],
    \label{eq:fourier}
\end{align}
where $f$ is an integrable function such that
\begin{equation}
  \begin{cases}
    f(\frac{n+\frac{1}{2}}{N}) = x_n, \\
    f(t) = f(-t), \\
    f(t) = 0, \quad t > 1,
    \label{eq:cond}
  \end{cases}
\end{equation}
and $F$ is the Fourier transform of $f$.  By this equation, if
sequence $x_0, \ldots, x_{N-1}$ consists of equally spaced samples of
function $f$ and $f$ satisfies Eq.~\eqref{eq:cond}, then a DCT
coefficient of the sequence is a sample of $f$ in frequency domain.

\subsection{Quantization Effects on Linear Signal}

Suppose input signal $\boldsymbol{x}_r=\{x_0, \ldots, x_{N-1}\}$ is a
decreasing linear sequence, in which the $n$-th element is,
\begin{equation}
  x_n = a \cdot \frac{n+\frac{1}{2}}{N}, \quad 0 \le n \le N-1.
\end{equation}
Then triangular function $a \cdot \operatorname{tri}(t)$, where,
\begin{align}
  \operatorname{tri}(t) =
    \begin{cases}
      1 - |t| & \mbox{if }|t|<1 \\
      0 & \mbox{otherwise}, \\
    \end{cases}
\end{align}
satisfies the conditions in Eq.~\eqref{eq:cond}, thus by
Eq.~\eqref{eq:fourier}, the $k$-th DCT coefficient of the sequence can
be approximated as,
\begin{align}
  X_k
  &\approx \operatorname{Re} \left[ \frac{N}{2}
    \int_{-\infty}^\infty a \cdot \operatorname{tri}(t)
    e^{-i 2 \pi t \frac{k}{2}} \D{t} \right] \nonumber
  \\
  &= \frac{aN}{2} \cdot \operatorname{sinc}^2
    \left( \frac{k}{2} \right),
    \label{eq:dctramp}
\end{align}
where $\operatorname{sinc}(\cdot)$ is the normalized sinc function
defined as,
\begin{equation}
  \operatorname{sinc}(x) = \frac{\sin(\pi x)}{\pi x}.
\end{equation}

\begin{figure}
  \centering

  \includegraphics[width=0.8\linewidth]{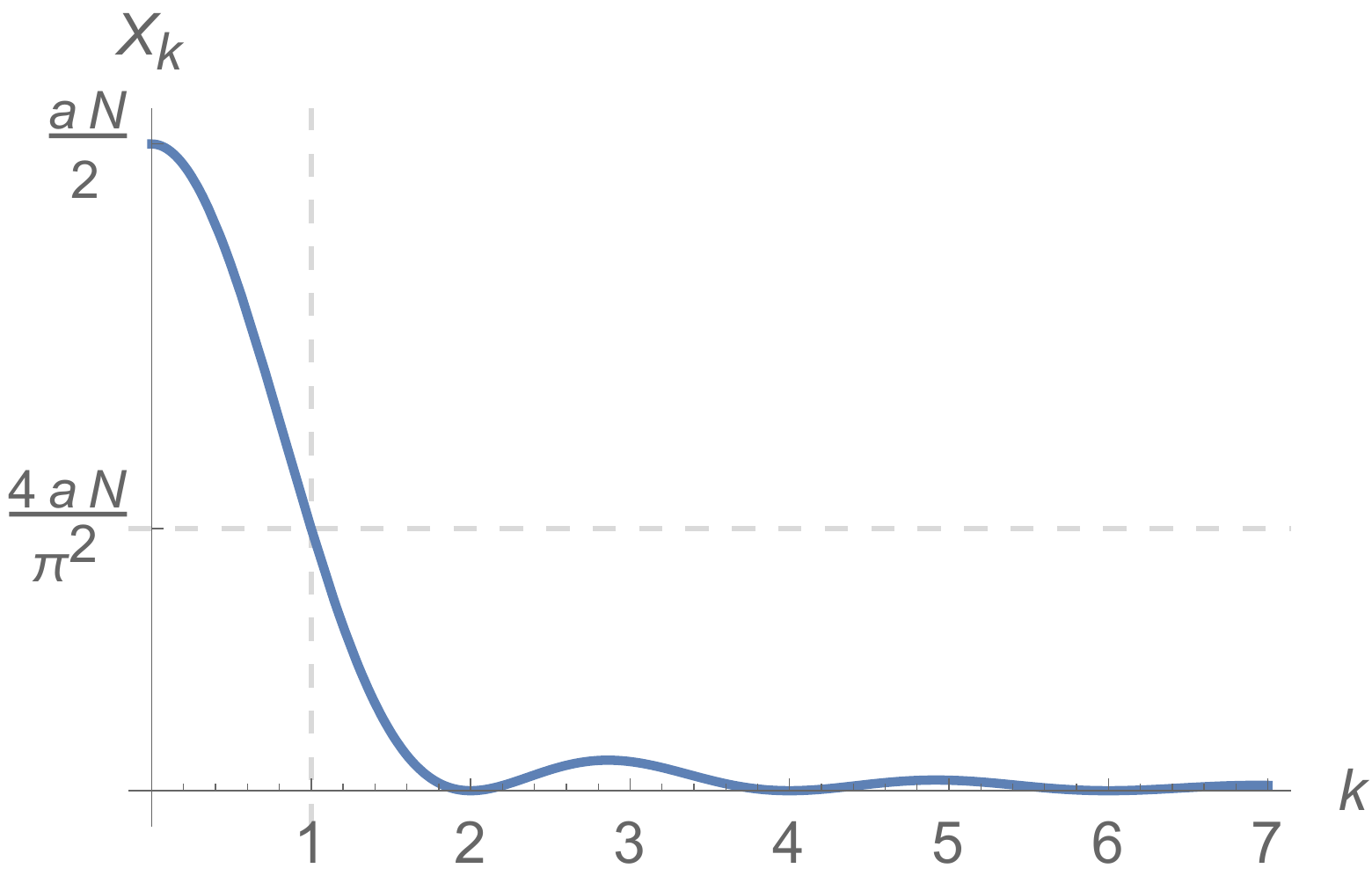}

  \caption{DCT coefficients of a decreasing linear sequence.}

  \label{fig:dctramp}
\end{figure}

\begin{figure}
  \centering

  \setlength{\subfigwidth}{0.45\linewidth}

  \begin{subfigure}[b]{\subfigwidth}
    \centering
    \includegraphics[width=\subfigwidth,
    height=0.6\subfigwidth,keepaspectratio=false]{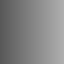}
    \caption{Original}
  \end{subfigure}
  \begin{subfigure}[b]{\subfigwidth}
    \centering
    \includegraphics[width=\subfigwidth,
    height=0.6\subfigwidth,keepaspectratio=false]{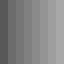}
    \caption{JPEG ($\mbox{QF}=10$)}
  \end{subfigure}

  \caption{Quantization removes small AC components causing
    perceivable blocking artifacts in a simple gradient ramp image.}

  \label{fig:jpegramp}
\end{figure}

This linear input signal $\boldsymbol{x}_r$ is easy to model in
temporal domain; its second order derivative is zero everywhere hence
sparse.  It is comparable to a simple gradient ramp in \gls{2d}
digital image.  However, as visualized in Fig.~\ref{fig:dctramp}, this
signal is not sparse in DCT domain.  By the approximation in
Eq.~\eqref{eq:dctramp}, the $k$-th DCT coefficient $X_k$ is only zero
for even positive integer $k$, thus, more than half ($\lfloor N/2
\rfloor+1$ out of $N$) of the coefficients in DCT domain are non-zero.
To effectively compress this signal, some of the non-zero DCT
coefficients need to be quantized to zero.  Suppose for some odd
positive integer $k_0$, the $k_0$-th quantized DCT coefficient is
zero, then
\begin{align}
  X_{k_0} < \frac{q_{k_0}}{2}
  & \Longleftrightarrow
  \frac{aN}{2} \cdot \operatorname{sinc}^2
    \left( \frac{k_0}{2} \right) < \frac{q_{k_0}}{2}
    \nonumber
  \\
  & \Longleftrightarrow
  aN \cdot \frac{4}{\pi^2 k_0^2} < q_{k_0}
    \nonumber
  \\
  & \Longleftrightarrow
  a < \frac{\pi^2 k_0^2 q_{k_0}}{4N}.
    \label{eq:bounda}
\end{align}
Since non-zero DCT coefficient $X_k$ decreases with $k$ while in
general, quantization interval $q_k$ increases with $k$, all the
quantized DCT coefficients after the $k_0$-th one are zero as well.
For example, if we use JPEG with $\mbox{QF}=25$ to compress a
horizontal gradient ramp, then by Eq.~\eqref{eq:bounda}, quantized DCT
coefficient $X_1$ is zero when $a$ is less than about $4.8$.  In this
case, each coding block becomes uniform after compression as its AC
components in DCT domain are all zeros.  As shown in
Fig.~\ref{fig:jpegramp}, it is not sufficient to reconstruct the
gradient ramp accurately in each block with only the DC component
$X_0$.  More importantly, due to Mach bands illusion, the
discontinuity around coding block boundaries is highly perceivable to
human, greatly deteriorating the perceptual quality of the compressed
image.

\subsection{Quantization Effects on Piecewise Constant Signal}

Similar to linear signal, piecewise constant signal is another case
which is simple to model in temporal domain but complex in DCT domain.
For instance, let input signal $\boldsymbol{x}_s = \{x_0, \ldots,
x_{N-1}\}$ be a sequence of two steps, i.e.,
\begin{equation}
  [\underbrace{a,a,\ldots,a}_{m},\underbrace{0,0,\ldots,0}_{N-m}],
\end{equation}
where $a>0$ and $0 \le m \le N$.  This sequence is a discrete version
of rectangular function $f_s(t)$, where,
\begin{align}
  f_s(t)
  &= a \cdot \operatorname{rect}\left(\frac{t}{2r} \right) \nonumber
  \\
  &= a \cdot
    \begin{cases}
      0 & \mbox{if }|t|>r \\
      \frac{1}{2} & \mbox{if }|t|=r \\
      1 & \mbox{if }|t|<r,
    \end{cases}
\end{align}
and $r=m/N$.  As $f_s(t)$ satisfies Eq.~\eqref{eq:cond}, the \gls{dct}
of sequence $\boldsymbol{x}_s$ can be approximated as follows by
Eq.~\eqref{eq:fourier},
\begin{align}
  X_k &\approx \operatorname{Re} \left[ \frac{N}{2}
        \int_{-\infty}^\infty a \cdot \operatorname{rect}\left(\frac{t}{2r} \right)
        e^{-i 2 \pi t \frac{k}{2}} \D{t} \right] \nonumber
  \\
  &= \frac{N}{2} \cdot 2ar \cdot \operatorname{sinc}
    \left( 2r \cdot \frac{k}{2} \right) \nonumber
  \\
  &= arN \cdot \operatorname{sinc}(rk).
\end{align}

As $\operatorname{sinc}(rk)$ decreases with frequency $k$ in general,
if quantization intervals are large enough, quantization effects can
be approximated by cutting off high frequency components.  Suppose
only the first $b$ DCT coefficients are preserved after quantization,
then the restored sequence is
\begin{align}
  \label{eq:restore}
  \hat{x}_t
  &= \int_{-b}^b \frac{2}{N} X_k \,
    e^{i 2 \pi t \frac{k}{2}} \D{\frac{k}{2}}
    \nonumber
  \\
  &= a \int_{-b}^b r \operatorname{sinc}(rk) \,
    e^{i 2 \pi \frac{t}{2} k} \D{k}
    \nonumber
  \\
  &= a \cdot [\operatorname{Si}(br-bt)+\operatorname{Si}(br+bt)],
\end{align}
where function $\operatorname{Si}(z)$ is sine integral defined as
\begin{equation}
  \operatorname{Si}(z) = \int_0^z \operatorname{sinc}(t) \D{t}.
\end{equation}
By aligning the sequence $\{\hat{x}_t \,|\, 0 \le t \le 1\}$ to the
location of the edge, we get the following sequence such that $y_0$ is
the transition of two steps,
\begin{equation}
  \hat{y}_t = \hat{x}_{t+r} = a [\operatorname{Si}(-bt)
  +\operatorname{Si}(2br+bt)].
\end{equation}

\begin{figure}
  \centering
  \includegraphics[width=0.8 \linewidth]{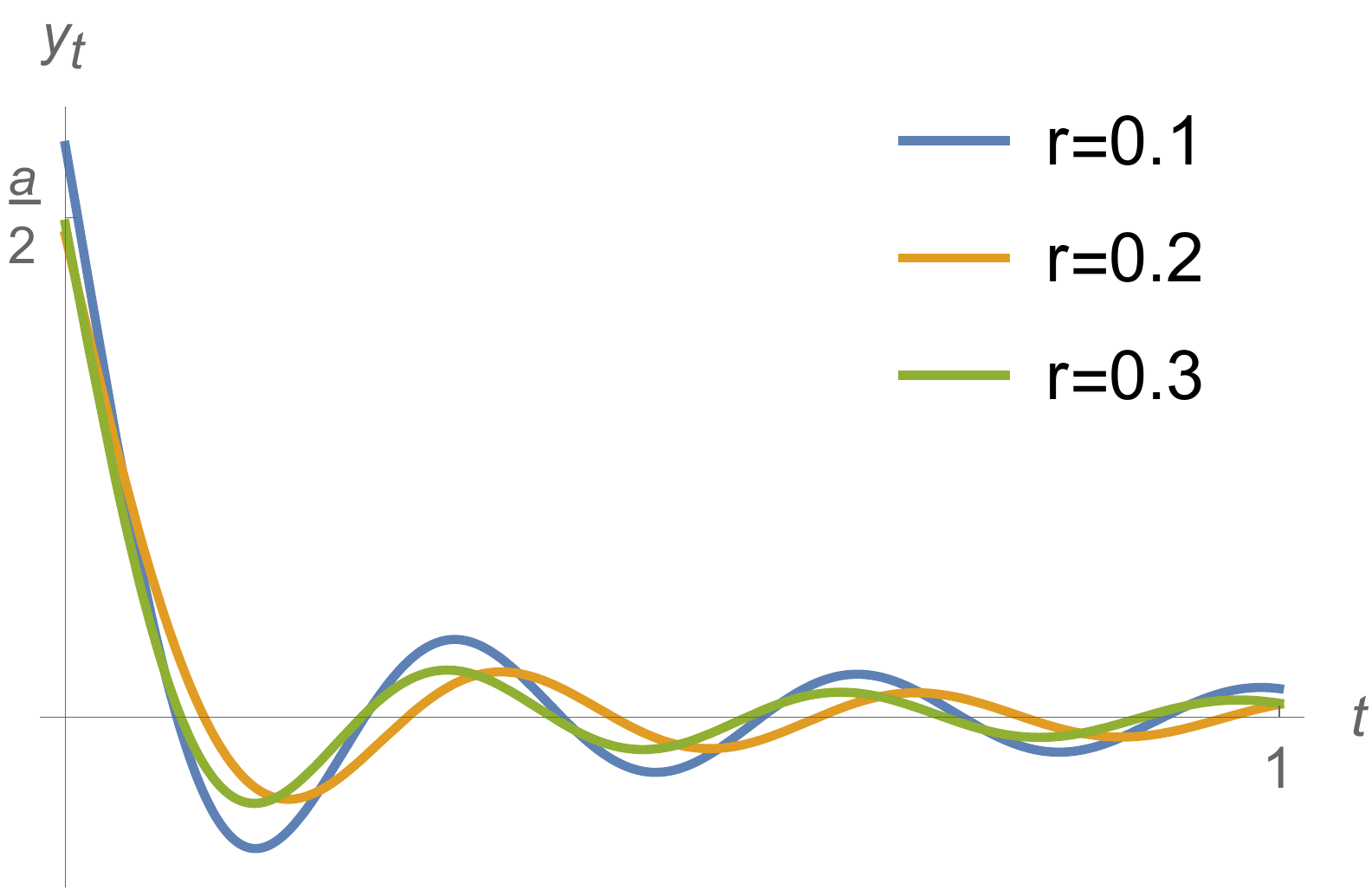}
  \caption{A sharp edge causes similar quantization artifacts
    regardless of the phase $r$.}
  \label{fig:restore}
\end{figure}

As shown in Fig.~\ref{fig:restore}, quantization noise in $\hat{y}_t$
has a relatively fixed pattern regardless of the phase.  Therefore, if
we align the signals by their phases, the noises become aligned as
well.  This correlation between signal and quantization noise makes
them much more difficult to distinguish.

If sequence $\boldsymbol{x}_s$ is smoothed with a Gaussian kernel
resulting sequence $\boldsymbol{y}_s$, then $\boldsymbol{y}_s$ is a
discrete version of $f_g(t) = (f_s*g)(t)$, where,
\begin{equation}
  g(t) = \frac{1}{\sigma \sqrt{2 \pi}} e^{-\frac{t^2}{2\sigma^2}}.
\end{equation}
By convolution theorem, the Fourier transform of function $f_g(t)$ is,
\begin{align}
  F_g\left(\frac{k}{2}\right)
  &= F_s\left(\frac{k}{2}\right)
    \cdot G\left(\frac{k}{2}\right) \nonumber
  \\
  &= 2ar \cdot \operatorname{sinc}(rk)
    \cdot e^{-2 \pi^2 \sigma^2 k^2 / 2^2}
\end{align}
Thus the $k$-th \gls{dct} coefficient of the blurred sequence
$\boldsymbol{y}_s$ is approximately equal to,
\begin{equation}
  Y_k \approx arN \cdot \operatorname{sinc}(rk)
  \cdot e^{- \pi^2 \sigma^2 k^2 / 2}
  \label{eq:dcty}
\end{equation}

For a given frequency $k$, the absolute quantization error is
\begin{equation}
  \epsilon_k = \left| v_k \cdot q_k - Y_k \right|.
\end{equation}
where $v_k= \lfloor Y_k/q_k + 0.5 \rfloor$.  If the absolute
quantization error $\epsilon_k$ of \gls{dct} coefficient $Y_k$ is
sufficiently small, say less than a constant $C_\epsilon$, it has
little impact on the quality of the compressed image; if $\epsilon_k$
is large, but the relative error $\epsilon_k / |Y_k|$ is small, it
still contributes little to the artifacts of the compressed image, as
in this case, the quantized \gls{dct} coefficient is strong enough to
hide the error perceptually.  Suppose that to hide the quantization
artifacts from frequency $k$, the relative error $\epsilon_k / |Y_k|$
must be less than $1/3$, i.e.,
\begin{equation}
  \frac{e_k}{|Y_k|} = \left| \frac{ v_k \cdot q_k - Y_k}{Y_k} \right|
  < \frac{1}{3}. \\
\end{equation}
This inequality is true if and only if $|Y_k| \ge 3q_k/4$.  Thus, when,
\begin{equation}
  C_\epsilon \le |Y_k| \le \frac{3}{4}q_k,
\end{equation}
quantization of DCT coefficient $Y_k$ results both large absolute
error and relative error.

\begin{figure}
  \centering




  \begin{subfigure}[b]{0.48\linewidth}
    \centering
    \includegraphics[width=\linewidth]{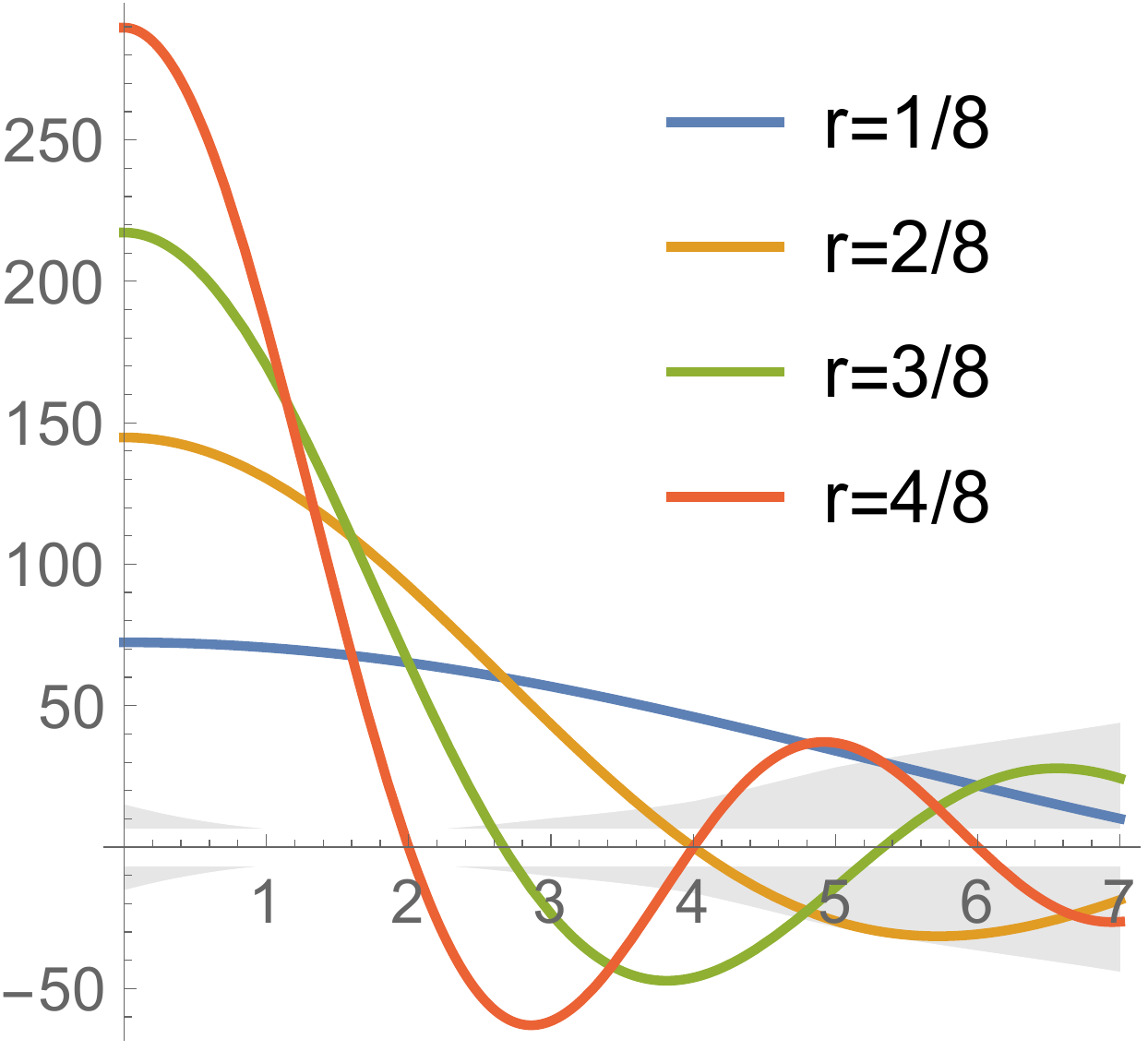}
    \caption{$\sigma=0$}
    \label{fig:dcte1}
  \end{subfigure}
  \begin{subfigure}[b]{0.48\linewidth}
    \centering
    \includegraphics[width=\linewidth]{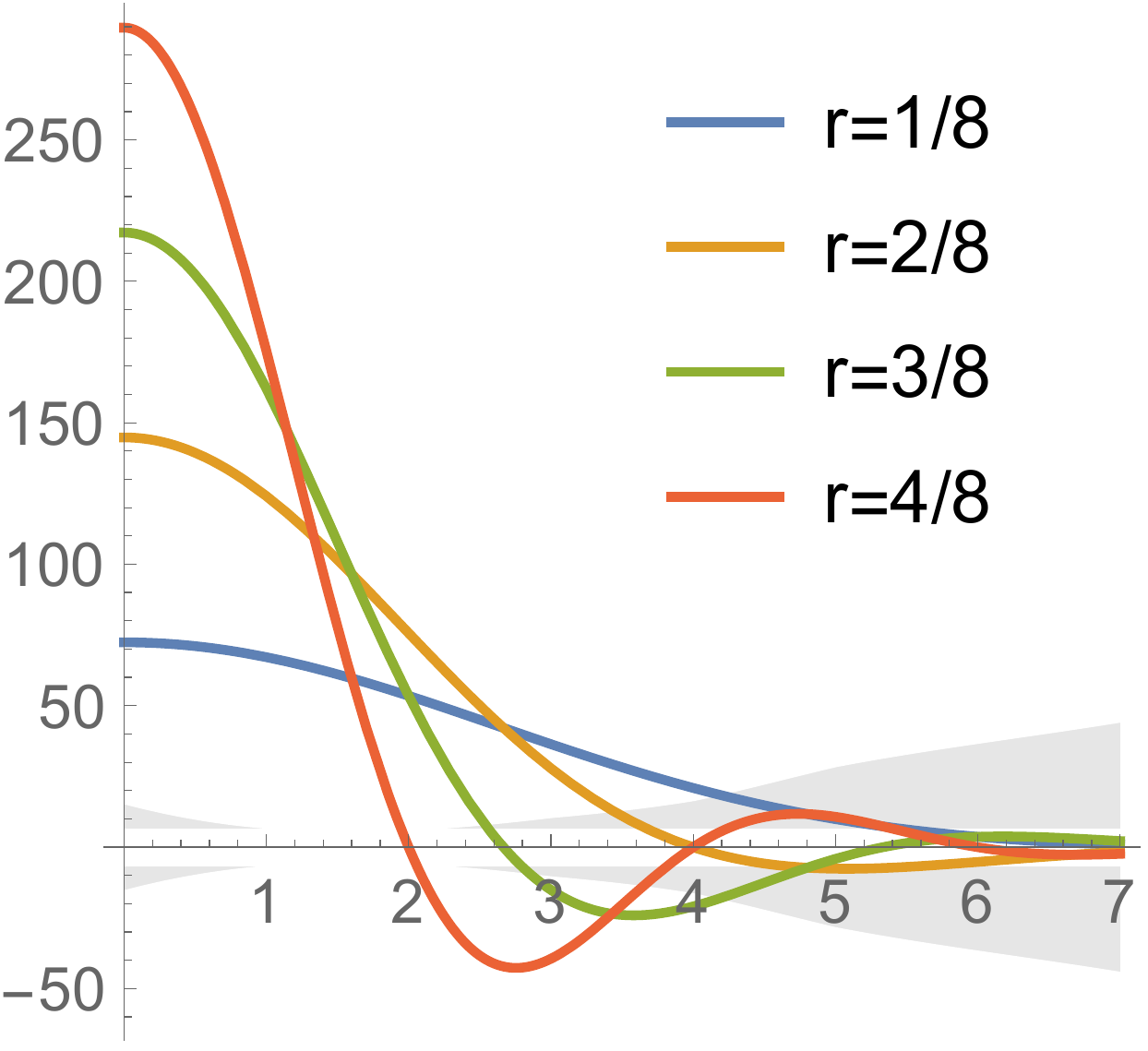}
    \caption{$\sigma=0.1$}
    \label{fig:dcte2}
  \end{subfigure}

  \caption{\gls{dct} coefficient $Y_k$ as a function of frequency
    $k$.  }
  \label{fig:dcte}
\end{figure}

Plotted in Fig.~\ref{fig:dcte} is /gls{dct} coefficient $Y_k$ as a
function of $k$.  Each curve represents a signal with a different
phase $r$, and amplitude of the first step is $a = 50$; and the
quantization intervals are based on the quantization matrix of JPEG
with $\mbox{QF}=50$.  Regions where a coefficient can cause large
absolute error and relative quantization error are marked as gray.  As
demonstrated in Fig.~\ref{fig:dcte1}, the strength and sign of a
/gls{dct} coefficient and its quantization error depend on various
factors, e.g., phase $r$, smoothness $\sigma$ and amplitude $a$.
In Fig.~\ref{fig:dcte2}, the sequence is smoothed by a Gaussian kernel
with variance $\sigma^2$.  As a result, the quantization errors of
high frequency coefficients are small (not in the gray regions)
compare to the previous case, as those coefficients are close to zero.

\section{Enhancement Model}

Recent \gls{nns} based image denoising techniques, such as BM3D
\cite{dabov2006image}, SAIST \cite{dong2013nonlocal} and WNNM
\cite{gu2014weighted}, have demonstrated their great strength in
reconstructing the original image $\boldsymbol{x}$ from an observation
$\boldsymbol{y} = \boldsymbol{x} + \boldsymbol{n}$ contaminated by
additive white Gaussian noise $\boldsymbol{n}$.
\gls{nns} refers to the fact that there are many repeated local patterns
across a natural image, and those nonlocal similar patches to a given
patch can help much the reconstruction of it \cite{cai2010singular}.
For a local patch $\boldsymbol{y}_i$ of size $m$ in image
$\boldsymbol{y}$, we can stack $M$ of its similar patches
$\boldsymbol{y}_{i,j} = \boldsymbol{R}_{i,j}y$ across the image
together into a matrix $\boldsymbol{Y}_i \in \Re^{m \times M}$,
where $\boldsymbol{R}_{i,j}$ is a matrix extracting the $j$-th similar
patch of the local patch at location $i$ for $1 \le i \le N, 1 \le j
\le M$.  Then solving the following nuclear norm minimization (NNM)
problem yields a matrix $\boldsymbol{\hat{X}}_i$ consisting of noise
reduced patches,
\begin{equation}
  \boldsymbol{\hat{X}}_i = \underset{\boldsymbol{X}_i}{\operatorname{argmin}}
  \|\boldsymbol{Y}_i - {\boldsymbol{X}_i}\|_F^2
  + \lambda\|\boldsymbol{X}_i\|_*
  \label{eq:nnm}
\end{equation}
Although this problem in non-convex, it is tractable by an efficient
\gls{svt} algorithm \cite{cai2010singular}.  The whole reconstructed
image can be then estimated by aggregating all the denoised patches
as,
\begin{equation}
  \boldsymbol{\hat{x}} = \left( \sum_{i=1}^N \sum_{j=1}^M
    \boldsymbol{R}_{i,j}^{\intercal} \boldsymbol{R}_{i,j} \right)^{-1}
  \sum_{i=1}^N \sum_{j=1}^M \boldsymbol{R}^{\intercal}_{i,j}
  \boldsymbol{\hat{x}}_{i,j}
  \label{eq:aggregate}
\end{equation}

Following this idea, we can formulate the restoration of
\gls{dct}-domain compressed image problem as a constrained nuclear
norm minimization problem,
\begin{equation}
  \begin{array}{rl}
    \boldsymbol{\hat{x}} = \underset{\boldsymbol{x}}{\operatorname{argmin}}
    & \displaystyle\sum^{N}_{i=1}\|\boldsymbol{X}_i\|_*
    \\[3ex]
    \operatorname{s.t.}
    & |\boldsymbol{QT\tilde{R}}_{i}\boldsymbol{Hx} - \boldsymbol{\gamma}_i| \le 0.5,
    \\
    & \hfill i = 1 \ldots n
  \end{array}
  \label{eq:model}
\end{equation}
where $\boldsymbol{H}$ is a matrix modelling the degradation of image
quality caused by various image capturing conditions,
$\boldsymbol{\tilde{R}}_{i}$ is a matrix extracting the coding block
at location $i$, $\boldsymbol{T}$ is the \gls{dct} transform matrix,
$\boldsymbol{Q}$ is a diagonal matrix storing quantization table and
vector $\boldsymbol{\gamma}_i$ is the \gls{dct} coefficient of the
block at location $i$.

Using hard-decoding technique, each block $\boldsymbol{y}_i$ of
$\boldsymbol{y}$ is obtained by inverse \gls{dct} transform from coefficient
vector $\boldsymbol{\gamma}_i$ as follow,
\begin{equation}
  \boldsymbol{y}_i = \boldsymbol{\tilde{R}}_i\boldsymbol{y} =
  (\boldsymbol{QT})^{-1} \boldsymbol{\gamma}_i,
  \label{eq:yi}
\end{equation}
and the observed image $\boldsymbol{y}$ is a degraded version of image
$\boldsymbol{Hx}$ contaminated mainly by \gls{dct}-domain quantization
noise.  Existing sparsity based denoising techniques designed for
reducing additive white Gaussian noise generally use
$\|\boldsymbol{\hat{x}} - \boldsymbol{y}\|_2$ as the fidelity term in
their optimization frameworks and leave $\boldsymbol{\hat{x}}$
unconstrained.  In the case of \gls{dct}-domain quantization noise, we have
more information about the noise: the true value of each \gls{dct}
coefficient before scalar quantization lies in a known interval,
\begin{equation}
  \boldsymbol{Q}^{-1}(\boldsymbol{\gamma}_i - 0.5)
  \le \boldsymbol{T\tilde{R}}_{i}\boldsymbol{Hx}
  < \boldsymbol{Q}^{-1}(\boldsymbol{\gamma}_i + 0.5)
  \label{eq:constraint}
\end{equation}
This constrain confines the solution space of the optimization problem
in Eq.~\eqref{eq:model}, preventing the sparsity objective function
from over-smoothing the output image.

To solve the problem in Eq.~\eqref{eq:model}, we split it into two
parts.  The first part is to find a sparse estimation
$\boldsymbol{\hat{z}}$ of the original image from a given noisy
version $\boldsymbol{\hat{x}}$, i.e., for each patch group of image
$\boldsymbol{\hat{z}}$,
\begin{equation}
  \boldsymbol{\hat{Z}}_i = \underset{\boldsymbol{Z}_i}{\operatorname{argmin}}
  \|\boldsymbol{\hat{X}}_i - \boldsymbol{Z}_i\|_F^2 + \lambda \|\boldsymbol{Z}_i\|_*
  \label{eq:cleanup}
\end{equation}
The noisy version $\boldsymbol{\hat{x}}$ of the original image can be
estimated directly using $\boldsymbol{\hat{x}} =
\boldsymbol{H}^{-1}\boldsymbol{y}$.  Here we use $\boldsymbol{H}^{-1}$
to represent an inverse operator of $\boldsymbol{H}$ rather than
matrix inverse.  Although many types of image degradation can be
modelled by a simple product of a degradation matrix $\boldsymbol{H}$
and the original image $\boldsymbol{x}$, the inverse problem is often
iso-posed and requires complex non-linear algorithm to find a good
solution.  Since the observed image $\boldsymbol{y}$ contains
compression noise, if operator $\boldsymbol{H}^{-1}$ exhibits
high-boosting property, which is often the case for unsharp and edge
enhancement operators, $\boldsymbol{H}^{-1}$ could amplify the noise
and make an inaccurate estimation of the original image.  Using
sparsity prior, the boosted noise can be greatly alleviated by the
optimization problem in Eq.~\eqref{eq:cleanup}, resulting a better
estimation of the original image in vector $\boldsymbol{\hat{z}}$.

The second part of the problem is to impose the \gls{dct}-domain constraint
in Eq.~\eqref{eq:model} on the noise reduced estimation
$\boldsymbol{\hat{z}}$ from the first part using the following
optimization problem,
\begin{equation}
  \begin{array}{rl}
    \boldsymbol{\hat{x}}' = \underset{\boldsymbol{x}}{\operatorname{argmin}}
    & \|\boldsymbol{x} - \boldsymbol{\hat{z}}\|_2
    \\
    \operatorname{s.t.}
    & |\boldsymbol{QT\tilde{R}}_{i}\boldsymbol{Hx}
      - \boldsymbol{\gamma}_i| \le 0.5,
    \\
    & \hfill i = 1 \ldots n
  \end{array}
\end{equation}
This is a convex problem solvable by off-the-shelf convex optimization
problem solvers.  However, if the input image is large, a general
purpose solver is too time-consuming for this problem.  Instead, we
can solve a similar but much simpler problem as follows,
\begin{equation}
  \begin{array}{rl}
    \boldsymbol{\hat{x}}' = \underset{\boldsymbol{x}}{\operatorname{argmin}}
    & \displaystyle\sum_{i=1}^n \|\boldsymbol{\tilde{R}}_{i}
      \boldsymbol{H}(\boldsymbol{x} - \boldsymbol{\hat{z}})\|_2
    \\[3ex]
    \operatorname{s.t.}
    & |\boldsymbol{QT\tilde{R}}_{i}\boldsymbol{Hx}
      - \boldsymbol{\gamma}_i| \le 0.5,\\
    & \hfill i = 1 \ldots n
  \end{array}
  \label{eq:yk}
\end{equation}
Compared with the original problem, the only difference of the reduced
problem is that the new problem measures the norm of the error in
degraded image domain rather than original image domain.  Since the
\gls{dct} tranform matrix $\boldsymbol{T}$ is unitary,
\begin{align}
  \|\boldsymbol{\tilde{R}}_{i}\boldsymbol{H}(\boldsymbol{x}
  - \boldsymbol{\hat{z}})\|_2
  & = \|\boldsymbol{T\tilde{R}}_{i}\boldsymbol{H}(\boldsymbol{x}
    - \boldsymbol{\hat{z}})\|_2 \nonumber
  \\
  & = \|\boldsymbol{Ty}_i
    - \boldsymbol{T\tilde{R}}_{i}\boldsymbol{H\hat{z}}\|_2,
\end{align}
where coding block $\boldsymbol{y}_i = \boldsymbol{\tilde{R}}_{i}
\boldsymbol{H} \boldsymbol{x}$.  On the other hand, the \gls{dct}
coefficient constraint in Eq.~\eqref{eq:constraint} is applied on each
element of vector $\boldsymbol{Ty}_i$, thus, the optimization problem
has a closed-form solution,
\begin{equation}
  \boldsymbol{\hat{y}}_i = \mathcal{C}_{0.5}(\boldsymbol{\tilde{R}}_{i}
  \boldsymbol{H\hat{z}}, \boldsymbol{y}_i),
  \label{eq:yi}
\end{equation}
where, the \gls{dct}-domain clipping operator $\mathcal{C}_\beta
(\cdot, \cdot)$ is defined as
\begin{align}
  \mathcal{C}_\beta ( \boldsymbol{\alpha}, \boldsymbol{\rho} )
  =
  (\boldsymbol{QT})^{-1} \min(
  \max(\boldsymbol{QT\alpha}&,
    \boldsymbol{QT\rho}-\beta) \nonumber
  \\
  &,\boldsymbol{QT\rho}+\beta).
  \label{eq:clip}
\end{align}
Aggregating these \gls{dct} blocks together, we get
$\boldsymbol{\hat{y}}$, an estimation in degraded image domain with
reduced compression noise, from which an approximate solution
$\boldsymbol{\hat{x}}' = \boldsymbol{H}^{-1} \boldsymbol{\hat{y}}$ of
the problem in Eq.~\eqref{eq:yk} can be easily found.  Compared with
image $\boldsymbol{\hat{x}}$, the initial inverse of the observed
image $\boldsymbol{y}$, $\boldsymbol{\hat{x}}'$ has lower level of
compression noise because of sparsity prior in Eq.~\eqref{eq:cleanup}
and still satisfies the \gls{dct} domain constraints due to
Eq.~\eqref{eq:yk}.

\section{Algorithm}

\begin{algorithm}
  \caption{Image restoration from compressed image}

  \textbf{Input:} Compressed image $\boldsymbol{y}$, contrast
  degradation matrix $\boldsymbol{H}$

  \begin{algorithmic}[1]
    \State Estimate compression RMSE $\varepsilon$ of $\boldsymbol{y}$
    \State Estimate threshold $\lambda$ using $\varepsilon$
    \Comment{Eq.~\eqref{eq:lambda}}
    \State $\beta = 0.2$
    \State $\boldsymbol{x}^{(0)} = \boldsymbol{H}^{-1}\boldsymbol{y}$
    \For {$k=1$ to $K$}
    \For {\textbf{each} patch $\boldsymbol{x}_i$ in $\boldsymbol{x}^{(k-1)}$}
    \State Find similar patch group $\boldsymbol{X}_i$
    \State $[\boldsymbol{U},\boldsymbol{\Sigma},
    \boldsymbol{V}]=\operatorname{SVD}(\boldsymbol{X}_i)$
    \State $\boldsymbol{Z}_i= \boldsymbol{U}\mathcal{T}_\lambda
    (\boldsymbol{\Sigma})\boldsymbol{V}^{\intercal}$ \Comment{Eq.~\eqref{eq:hardsvt}}
    \EndFor
    \State Aggregate $\boldsymbol{Z}_i$ to form image $\boldsymbol{z}^{(k)}$ \Comment{Eq.~\eqref{eq:aggregate}}
    \If {$k = K$}
    \State $\beta = 0.5$
    \EndIf
    \State Clip $\boldsymbol{z}^{(k)}$ using threshold $\beta$ to get $\boldsymbol{y}^{(k)}$ \Comment{Eq.~\eqref{eq:yi}}
    \State $\boldsymbol{x}^{(k)} = \boldsymbol{H}^{-1}\boldsymbol{y}^{(k)}$
    \State $\lambda = \lambda / 2$
    \EndFor
  \end{algorithmic}

  \textbf{Output:} Restored image $\boldsymbol{x}^{(K)}$
  \label{alg:ours}
\end{algorithm}

Based on the restoration model discussed in the previous section, our
purposed algorithm can be implemented as an iterative process
alternatively finding a reconstructed original image and a compression
noise reduced observation image as in Algorithm \ref{alg:ours}.  In
this section, we address some of the technical issues in the
implementation of the algorithm.

\subsection{Nonlocal Self-Similarity}

\Gls{nns} based techniques have achieved the state-of-the-art results
in removing Gaussian noise \cite{gu2014weighted}.  The \gls{nns} prior
assumes that noise is independent to signal, hence by comparing a
group of similar patches, noise can be isolated from signal.  However,
as we argued in previous sections, compression noise is not random but
correlated with the signal; similar patches have similar compression
noise, especially when they also have the same relative position to
DCT coding blocks.  Moreover, patches with matched artifacts can be
easily mistaken as being similar using square error metric.  Thus,
collecting similar patches without taking their contents or positions
into consideration inevitably puts multiple instances of the same
quantization artifacts into a sample patch group; consequently, such
reoccurring noises cannot be separated from the true signal by the
\gls{nns} prior alone.

\begin{figure}
  \centering
  \includegraphics[width=0.6 \linewidth]{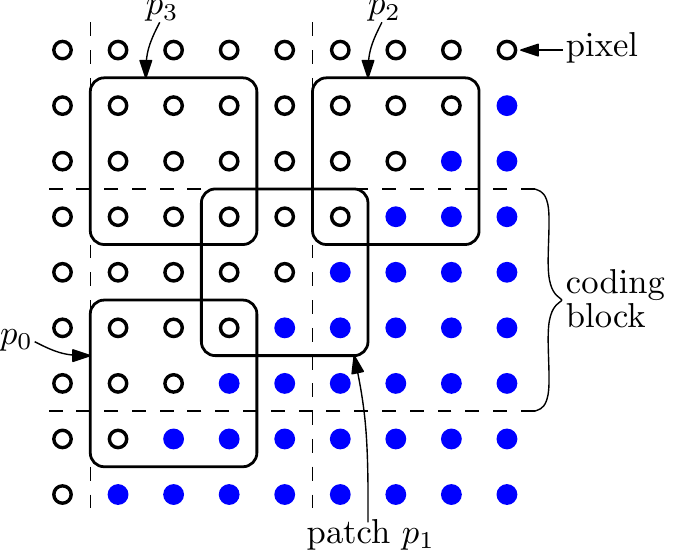}
  \caption{Avoid collecting patches of the same phase.}
  \label{fig:phase}
\end{figure}

For example, as shown in Fig.~\ref{fig:phase}, patches $p_0$ and $p_2$
are both located on a $45^{\circ}$ high-contrast edge, and their
positions relative to coding blocks are the same, hence they have
matching ringing artifacts caused by the quantization of the edge in
\gls{dct} domain.  In contrast, patch $p_1$ on the same edge also
suffers from ringing artifacts but with a different pattern than those
of $p_0$ and $p_1$ as a result of being aligned differently to coding
blocks than the other two.  Due to its distinct noise patterns, patch
$p_1$ is ranked lower in terms of the similarity to $p_0$, however, it
is a better candidate for the sample patch group in combating
reoccurring artifacts.  Therefore, when compiling the sample patch
group for patch $p_0$, other patches of the same position in relative
to coding blocks, like $p_2$, should be avoided if $p_0$ is around a
high-contrast edge.  A special case is that, when the edge is
horizontal or vertical, patches in the same row or column are
potentially distorted by the same artifacts, thus their similarity
rating must be reduced accordingly as well.

In addition to considering patch positions in choosing similar
patches, the measurement of patch similarity should be carefully
designed to decouple noise from signal.  As discussed previously,
compression noise in input image can mislead the selection of similar
patches.  Thus, instead of comparing two patches directly by their
squared error, a denoised version of a patch can be used to measure
similarity.  A simple low-pass filtering can generate a good enough
denoised image effectively reducing the impact of compression noise to
the measurement of similarity.  This technique is only used in the
first iteration of a denosing algorithm when the strength of noise is
high.  In the later iterations, input image becomes less subject to
compression noise and it is not necessary to clean up the input image
for a more robust measurement of similarity.

The above discussed techniques are designed to deal with patches with
ringing artifacts around strong edges.  Patches in smooth areas are
generally free of ringing artifacts since their high frequency
coefficients are near zero and the corresponding quantization errors
are negligible.  However, these patches are not immune to blocking
artifacts.  Due to the lack of other textures, the boundaries of
coding blocks are actually more discernible perceptually in those
smooth areas as demonstrated in Fig.~\ref{fig:jpegramp}.  To prevent
these blocking artifacts being matched as similar patch features
causing reoccurring artifacts in sample patch group, the same strategy
of choosing only unaligned patches as previous case can be employed.
For example, in Fig.~\ref{fig:phase}, patch $p_3$ is in a smooth area
located across two coding blocks; any patch in the same row as $p_3$
is likely to have the identical blocking artifacts, hence it should
not be considered in the patch group of $p_3$.  Moreover, since
natural images are smooth in general, patches in a small windows of
smooth area are similar to each other.  Furthermore, since natural
images are smooth in general, patches in a small windows of smooth
area are similar to each other.  Therefore, for a patch from a smooth
area, patches in close proximity are sufficient to build a good sample
patch group.  If the search window is small enough, there are few
patches perfectly aligned with the given patch, hence reducing the
risk of collecting too many patches with repeated artifacts.  In
practice, we set the search windows to $60 \times 60$ for normal patch
and shrink the window to $10 \times 10$ when the variance of the given
patch is less than $3$.

\subsection{Singular Value Thresholding}

Ideally, finding a low-rank reconstruction of patch group matrix
$\boldsymbol{Y}_i$ should be formulated as an $\ell_0$-norm
minimization problem as follows,
\begin{equation}
  \boldsymbol{\hat{X}}_i
  = \underset{\boldsymbol{X}_i}{\operatorname{argmin}}
  \|\boldsymbol{Y}_i - {\boldsymbol{X}_i}\|_F^2
  + \lambda\|\boldsymbol{X}_i\|_0 .
  \label{eq:l0m}
\end{equation}
Since this problem is NP-hard \cite{cai2010singular}, in practice, we
approximate it with a nuclear norm minimization problem as in
Eq.~\eqref{eq:nnm}, which has an efficient closed-form solution,
\begin{equation}
  \boldsymbol{\hat{X}}_i= \boldsymbol{U}\mathcal{D}_\lambda
  (\boldsymbol{\Sigma})\boldsymbol{V}^{\intercal},
  \label{eq:solution}
\end{equation}
where $\boldsymbol{U},\boldsymbol{\Sigma},\boldsymbol{V}$ represent
the the singular value decomposition (SVD) of $\boldsymbol{Y}_i$ and
$\mathcal{D}_\lambda(\cdot)$ is a soft-thresholding operator,
\begin{equation}
  \mathcal{D}_\lambda(\boldsymbol{\Sigma})_{jj} =
  \begin{cases}
     \boldsymbol{\Sigma}_{jj}-\lambda
     & \boldsymbol{\Sigma}_{jj} > \lambda,
     \\
     0 & \mbox{otherwise},
  \end{cases}
\end{equation}
or simply
$\mathcal{D}_\lambda(\boldsymbol{\Sigma})_{jj} = \max
(\boldsymbol{\Sigma}_{jj} - \lambda, 0)$.

Although this is a reasonable approximation employed by many
applications \cite{cai2010singular, xie2014optimal, dong2013nonlocal},
it still has some weaknesses.  One of its problems is that, in
addition to having a lower rank, the Frobenius norm of the optimal
solution $\boldsymbol{\hat{X}}_i$ also decreases with larger threshold
$\lambda$, since,
\begin{equation}
  \|{\boldsymbol{\hat{X}}_i}\|_F^2
  =  \|\boldsymbol{U}\mathcal{D}_\lambda(\boldsymbol{\Sigma})
  \boldsymbol{V}^{\intercal}\|_F^2
  = \sum_{j=1}^M \mathcal{D}_\lambda(\boldsymbol{\Sigma})_{jj}^2
\end{equation}
and $\mathcal{D}_\lambda(\boldsymbol{\Sigma})_{jj}$ is a decreasing
function to $\lambda$.  In the context of image denoising, when we try
to increase the strength of the denoising algorithm by selecting a
large threshold $\lambda$, it inevitably decreases the second moment
of the image reducing the brightness and contrast of the output.
Unlike white Gaussian noise, \gls{dct}-domain quantization noise could
contribute negatively to the second moment of the image, especially
when the quality factor is low, hence, image denoising using NNM may
pull the result further away from the statistics of the original
image.

An intuitive solution to this problem is to completely preserve all
the singular values that are above the threshold $\lambda$, i.e., to
replace the soft-thresholding operator $\mathcal{D}_\lambda(\cdot)$
with a hard-thresholding operator
\begin{equation}
  \mathcal{T}_\lambda(\boldsymbol{\Sigma})_{jj} =
  \begin{cases}
     \boldsymbol{\Sigma}_{jj} & \boldsymbol{\Sigma}_{jj} > \lambda,
     \\
     0 & \mbox{otherwise}.
  \end{cases}
  \label{eq:hardsvt}
\end{equation}
Since $\mathcal{T}_\lambda(\boldsymbol{\Sigma})_{jj} > 0$ if and only
if $\mathcal{T}_\lambda(\boldsymbol{\Sigma})_{jj} > 0$, the resulting
matrices $\boldsymbol{\hat{X}}_i$ by the two threshold operators have
the exact same rank.  Thus, the new operator $\mathcal{T}_\lambda
(\cdot)$ does not change the low rank property of the solution,
however, in this case, the solution is closer to $\boldsymbol{Y}_i$
statistically in terms of the second moment.

This method coincides with the idea of reweighted nuclear norm
minimization where large singular values are given smaller weight to
achieve better low rank approximation \cite{li2014reweighted}.  It can
also be interpreted as a spacial case of weighted nuclear norm
minimization (WNNM) \cite{gu2014weighted} as follows.  If for
$\sigma_j(\boldsymbol{X}_i)$, the $j$-th singular value of
$\boldsymbol{X}_i$, we assign a weight $w_j$,
\begin{equation}
  w_j =
  \begin{cases}
     0 & \sigma_j(\boldsymbol{Y}_i) > \lambda,
     \\
     \lambda & \mbox{otherwise}.
  \end{cases}
\end{equation}
Since the weights $w_{1 \ldots M}$ are in a non-descending order, by
the theory of WNNM, applying the hard-thresholding operator
$\mathcal{T}_\lambda(\cdot)$ on $\boldsymbol{\Sigma}$ yields an
optimal solution for optimization problem,
\begin{equation}
  \boldsymbol{\hat{X}}_i
  = \underset{\boldsymbol{X}_i}{\operatorname{argmin}}
  \|\boldsymbol{Y}_i - {\boldsymbol{X}_i}\|_F^2
  + \|\boldsymbol{X}_i\|_{\boldsymbol{w},*}
\end{equation}
where $\|\boldsymbol{X}_i\|_{\boldsymbol{w},*}$ is the weighted sum of
the singular values of matrix $\boldsymbol{X}_i$.

Now, the question is how to set the parameter $\lambda$ of the NNM
problem making it more effective against \gls{dct}-domain quantization
noise.  In the formulation of Eq.~\eqref{eq:nnm}, $\lambda$ is a
weight balancing the sparse and fidelity regularization terms.  If
sparsity is given too much weight, it tends to over-smooth the image
and cause degradation in brightness and contrast as discussed
previously; if the weight is too small, noise remains visible.  From
the perspective of the solution to the problem in
Eq.~\eqref{eq:solution}, $\lambda$ is a threshold eliminating small
singular values of matrix $\boldsymbol{Y}_i = \boldsymbol{U \Sigma
  V}^{\intercal}$, where row vector $\sigma_j
\boldsymbol{v}_j^{\intercal}$ in matrix $\boldsymbol{\Sigma
  V}^{\intercal}$ consists of the coefficient of each patch in
$\boldsymbol{Y}_i$ with respect to the $j$-th basis vector in sparse
dictionary $\boldsymbol{U}$ \cite{dong2013nonlocal}.  Image denoising
by sparse optimization is based on the fact that signal is likely
sparse under some basis while noise is i.i.d.~under the same basis.
Furthermore, the energy of compression error generally is small in
comparison with the strength of signal, especially when the
quantization factor is set to a practical range.  Thus, removing small
coefficients, which originated most likely from noise than signal,
results an output closer to the true signal.  The \gls{mse}
$\varepsilon^2$ of compression can then be approximated by,
\begin{align}
  \varepsilon^2 =
  & \frac{1}{mM} \|\boldsymbol{Y}_i - \boldsymbol{X}_i\|_F^2
    \nonumber \\
  \approx
  & \frac{1}{mM} \|\boldsymbol{Y}_i - \boldsymbol{\hat{X}}_i\|_F^2
    \nonumber \\
  =
  & \frac{1}{mM} \| \boldsymbol{U \Sigma V}^{\intercal} - \boldsymbol{U}
    \mathcal{T}_\lambda(\boldsymbol{\Sigma}) \boldsymbol{V}^{\intercal} \|_F^2
    \nonumber \\
  =
  & \frac{1}{mM} \| \boldsymbol{\Sigma}
    - \mathcal{T}_\lambda(\boldsymbol{\Sigma}) \|_F^2
    \nonumber \\
  =
  & \frac{1}{mM} \sum_{j \in L} \sigma_j^2
    \label{eq:mse}
\end{align}
where set $L$ contains indices of singular values that are less than
$\lambda$.  For the same input image, using a lower quality setting
increases the compression noise (i.e., $\varepsilon^2$), which in turn
increases the small singular values according to Eq.~\eqref{eq:mse}.
To compensate this, threshold $\lambda$ must increase as well to keep
the size of set $L$ unchanged so that the sparsity of the signal is
preserved.  Considering that the compression error is i.i.d.~on each
basis vector, implying that small singular values are of similar
strength, threshold $\lambda$ should be proportional to \gls{rmse}
$\varepsilon$ as,
\begin{equation}
  \varepsilon^2 \propto \frac{1}{mM} \sum_{j \in L} \lambda^2
\end{equation}
Therefore, with an empirical constant $C_\lambda$, threshold $\lambda$
can be set as
\begin{equation}
  \lambda = C_\lambda\varepsilon \sqrt{\frac{mM}{|L|}}
  \approx C_\lambda\varepsilon \sqrt{\max(M,m)},
  \label{eq:lambda}
\end{equation}
where we assume $|L| \approx \min(M,m)$ due to the sparsity of the
true signal.  This threshold selecting method requires the knowledge
of the strength of the compression error $\varepsilon$, which is
commonly unknown to the decoder.  If error $\varepsilon$ is indeed not
provided by the encoder, various no-reference \gls{psnr} estimation
techniques \cite{turaga2004no, ichigaya2006method, brandao2008no} can
be used to estimate $\varepsilon$ with sufficient accuracy for finding
an appropriate threshold $\lambda$.

\subsection{DCT Coefficient Constraint}

Most iterative denoising techniques, such as \cite{osher2005iterative,
  dong2013nonlocal, gu2014weighted}, employ some regularization
mechanisms to add a portion of filtered noise back to the denoised
image in each iteration in order to reduce the loss of high frequency
information as the result of multiple rounds of smoothing operators.
The idea of adding noise back enables denoising techniques to remove
large noise aggressively by over-smoothing the image in the first few
iterations without completely removing the detail in the process.
Then, during the following iterations, the image can be refined
gradually using smoothing operators of lower strength.  Mainly
designed to deal with Gaussian noise, many of these above mentioned
denoising techniques implement this iterative regularization by simply
adding the difference between the observed noisy image and smoothed
image back to the smoothed image, and use the result as the input
noisy image for the next iteration.

For our \gls{dct} quantization noise reduction algorithm, this noise
feedback method can be written as,
\begin{align}
  \boldsymbol{y}_i^{(k)}
  & = \boldsymbol{z}_i^{(k)}
    + \delta (\boldsymbol{y}_i - \boldsymbol{z}_i^{(k)})
    \nonumber
  \\
  & = \delta \boldsymbol{y}_i + (1 - \delta) \boldsymbol{z}_i^{(k)}
    \nonumber
  \\
  & = \boldsymbol{T}^{-1} [ \delta \boldsymbol{Ty}_i
    + (1 - \delta) \boldsymbol{Tz}_i^{(k)} ]
  \label{eq:feedback}
\end{align}
where coding block $\boldsymbol{y}_i$ is at location $i$ in the
observed image $\boldsymbol{y}$ as defined in Eq.~\eqref{eq:yi},
coding block $\boldsymbol{y}_i^{(k)}$ is the denoised version of
$\boldsymbol{y}_i$ from the $k$-the iteration of the algorithm, coding
block $\boldsymbol{z}_i^{(k)} = \boldsymbol{\tilde{R}}_i
\boldsymbol{Hz}^{(k)}$ is the smoothed block at location $i$ as in
Eq.~\eqref{eq:yk} and $\delta$ is a weight parameter adjusting the
strength of noise feedback.  As shown in Eq.~\ref{eq:feedback}, in
\gls{dct} domain, this noise feedback process finds a weighted average
between \gls{dct} coefficients $\boldsymbol{Tz}_i^{(k)}$ of the
smoothed image and coefficients $\boldsymbol{Ty}_i$ of the noisy
observation, adding image detail along with some reduced noise back to
the result.

Similarly, the clipping operator $\mathcal{C}_\beta
(\boldsymbol{\alpha}, \boldsymbol{\rho})$ introduced in the previous
section in Eq.~\eqref{eq:clip} has the effect of blending the smoothed
image $\boldsymbol{\alpha}$ with \gls{dct}-domain quantization error
tainted observation image $\boldsymbol{\rho}$ as well.  By design, the
clipping operator finds an image that is close to image
$\boldsymbol{\alpha}$ and has all of its \gls{dct} coefficients lying
within the given quantization intervals $[\boldsymbol{QT\rho}-\beta,
\boldsymbol{QT\rho}+\beta]$.  The output image $\mathcal{C}_\beta
(\boldsymbol{\alpha}, \boldsymbol{\rho})$ is closer to
$\boldsymbol{\alpha}$ if threshold $\beta$ is large, and it is closer
to $\boldsymbol{\rho}$ if $\beta$ is small.  Therefore, with
adjustable strength using parameter $\beta$, the clipping operator is
also a suitable noise feedback function for our algorithm as follows,
\begin{equation}
  \boldsymbol{y}_i^{(k)}
  = \mathcal{C}_\beta(\boldsymbol{z}_i^{(k)}, \boldsymbol{y}_i).
\end{equation}
This formulation is the same as the solution to the \gls{dct}
coefficient constraint problem in Eq.~\eqref{eq:yi} except for the
threshold $\beta$.  Since applying the clipping operator multiple
times is equivalent to applying it once with the smallest threshold
$\beta$, i.e.,
\begin{equation}
  \mathcal{C}_\beta(\mathcal{C}_{0.5}(\boldsymbol{z}_i^{(k)},
  \boldsymbol{y}_i), \boldsymbol{y}_i)
  = \mathcal{C}_{\min(\beta,0.5)}(\boldsymbol{z}_i^{(k)},
  \boldsymbol{y}_i),
\end{equation}
we only need to use the clipping operator once with threshold $\beta
\le 0.5$ to solve the \gls{dct} coefficient constraint problem and add
filtered noise back to the result.

By the theory of \gls{nqcs}, the \gls{dct} coefficient clipping
threshold $\beta$ should be sufficiently small in order to achieve the
optimal results in terms of \gls{psnr} \cite{park1999theory}.  For
example, the authors of \gls{nqcs} demonstrated that fixing threshold
$\beta=0.1$ is good enough for various images; several research papers
on JPEG image deblocking and denoising reported that setting
$\beta=0.3$ often yields best results \cite{zhai2008efficient,
  liew2004blocking, sun2007postprocessing}.  The best choice of
$\beta$ depends on the distributions of the \gls{dct} coefficients of
the original image, quantization factors and characteristics of the
smoothing technique.  Although smaller threshold $\beta$ generates
\gls{psnr}-plausible results, it often brings blocking and ringing
artifacts back to the result, deteriorating its perceptual visual
quality.

To alleviate this problem, in the last iteration of our algorithm,
instead of solving the optimization problem in Eq.~\eqref{eq:yk},
whose solution is given in Eq.~\eqref{eq:yi} using the clipping
operator, we solve a modified problem as follows,
\begin{equation}
  \begin{array}{rl}
    \boldsymbol{y}^{(k)} =
    \underset{\boldsymbol{y}}{\operatorname{argmin}}
    &  \| \nabla^2(\boldsymbol{y}
      -\boldsymbol{Hz}^{(k)}) \|_2^2
    \\[-1ex]
    & + \alpha \displaystyle \sum_{i=1}^n \|\boldsymbol{\tilde{R}}_{i}
      (\boldsymbol{y} - \boldsymbol{Hz}^{(k)})\|_2^2
    \\[3ex]
    \operatorname{s.t.}
    & |\boldsymbol{QT\tilde{R}}_{i}\boldsymbol{y}
      - \boldsymbol{\gamma}_i| \le \beta,
    \\
    & \hfill i = 1 \ldots n.
  \end{array}
  \label{eq:yk2}
\end{equation}
In addition to minimizing the difference between the smoothed image
and output image, the objective function of this modified problem also
minimizes the difference between their second order derivatives.  This
new regularization term encourages adding filtered noise back to
locations that are discontinuous in the smoothed image, so that,
artifacts are less noticeable in the output image perceptually.  The
modified problem in Eq.~\eqref{eq:yk2} is solvable using augmented
Lagrangian method, which is more expensive than the clipping operator
in Eq.~\eqref{eq:yi} in terms of computational complexity.  However,
since our algorithm only solves this problem once during the last
iteration, this technique can improve the visual quality of the output
image without significantly increasing the overall cost.

Alternatively, we can obtain the goal of eliminating the visual
artifacts by adjusting \gls{dct} clipping threshold $\beta$ and
singular value threshold $\lambda$ in the last two iterations.  The
idea is that, if in the last iteration $K$, most of the \gls{dct}
coefficients of smoothed block $\boldsymbol{z}_i^{(K)}$ are already
within the quantization intervals $[\boldsymbol{QTy}_i - 0.5,
\boldsymbol{QTy}_i + 0.5]$, then artifacts cannot be reintroduced to
the results by the clipping operator in Eq.~\eqref{eq:yi} with
threshold $\beta^{(K)} = 0.5$.  To insure the condition that most
\gls{dct} coefficients satisfy the quantization interval constraints,
the strength of the smoothing operator must be reduced in the last
iteration by using a smaller singular value threshold $\lambda^{(K)}$.
By Eqs.~\eqref{eq:mse} and \eqref{eq:lambda}, the standard deviation
of the difference between the noisy input image $\boldsymbol{y}$ and
smoothed image $\boldsymbol{z}^{(K)}$ is,
\begin{align}
  \frac{\| \boldsymbol{y} - \boldsymbol{z}^{(K)} \|_2}{\sqrt{n}}
  \approx \frac{\lambda^{(K)}}{C_\lambda\sqrt{\max(M,m)}},
\end{align}
which is roughly proportional to threshold $\lambda^{(K)}$, thus,
decreasing threshold $\lambda^{(K)}$ also reduces the variance of
$\boldsymbol{y} - \boldsymbol{z}^{(K)}$ in \gls{dct} domain and makes
\gls{dct} coefficients of image $\boldsymbol{z}^{(K)}$ more likely
stay within quantization interval.  On the other hand, the clipping
threshold $\beta^{(K-1)}$ in the second last iteration should also be
small in order to make each \gls{dct} coefficient of the clipped image
close to the centre of quantization interval, limiting \gls{dct}
coefficient overflow caused by the next smoothing operator.  However,
if both thresholds $\beta$ and $\lambda$ are too small, it weakens the
effect of noise reduction.  In practice, we find that setting clipping
threshold $\beta^{(K-1)} = 0.2$ and singular value threshold
$\lambda^{(K)} = \lambda^{(1)}/4$ works well for most input images.

\section{Experimental Results}

\begin{figure*}
  \centering

  \includegraphics[width=0.98\textwidth]{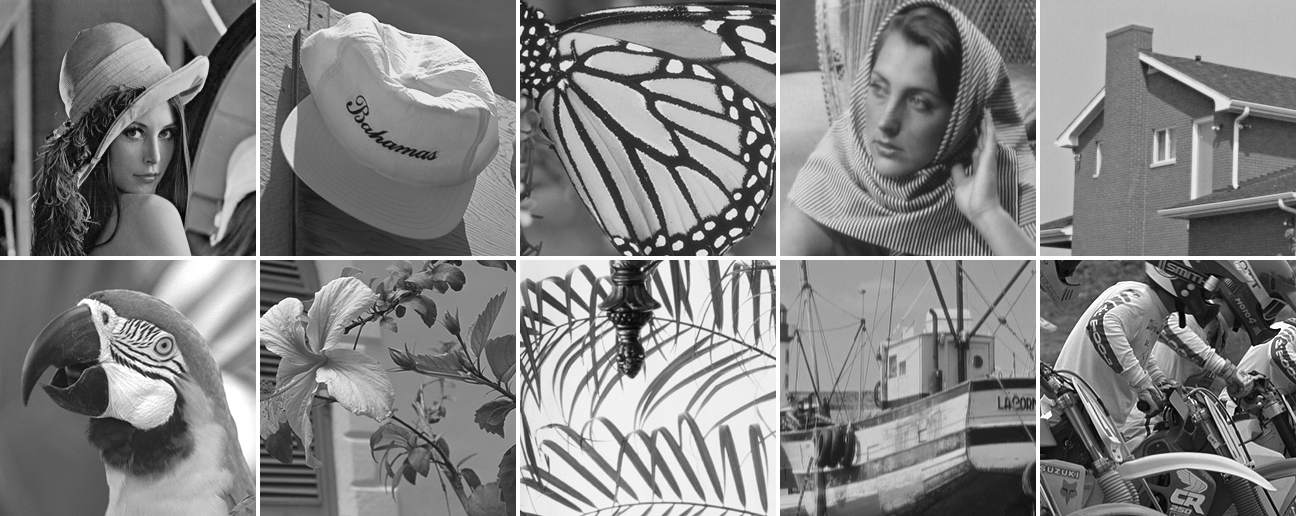}

  \caption{Several widely used test images.}

  \label{fig:testthumb}
\end{figure*}

To demonstrate the performance of the proposed technique, we first
turn off the image restoration part by setting degradation matrix
$\boldsymbol{H}$ as an identity matrix, and compare the results with
the state-of-the-art denoising and JPEG artifact removal techniques.
The comparison group is composed of the following methods: one JPEG
deblocking method: the ACR algorithm \cite{zhai2008efficient2}; two
denoising methods: the BM3D algorithm \cite{dabov2006image} and WNNM
algorithm \cite{gu2014weighted}; and three JPEG soft-decoding methods:
the TV algorithm \cite{bredies2012total}, DicTV
\cite{chang2014reducing} algorithm and DTPD algorithm
\cite{liu2015data}.  As the denoising approaches BM3D and WNNM are not
designed specifically for dealing with JPEG compression noise, they
cannot estimate the compression error from the input JPEG image but
require an estimation of the error variance as a user input.  To make
a fair comparison, we provide the true variance of the compression
error to these methods as a known parameter, so their performances
should reflect their best results in removing JPEG compression noise.

\begin{table}
\centering\footnotesize\begin{tabular}{|c|c|c|c|c|c|c|c|c|}
\hline
Image & JPEG & ACR & BM3D & WNNM & TV & DicTV & DTPD & Proposed \\
\hline\hline
Lenna & $30.64$ & $+0.23$ & $+0.92$ & $+0.66$ & $+0.12$ & $+0.48$ & $+1.45$ & $\mathbf{+1.58}$ \\
Parrot & $32.37$ & $+0.28$ & $+0.75$ & $-2.85$ & $-0.17$ & $+0.41$ & $+1.36$ & $\mathbf{+1.68}$ \\
Hat & $31.47$ & $+0.22$ & $+0.89$ & $-2.56$ & $+0.21$ & $+0.45$ & $+1.39$ & $\mathbf{+1.53}$ \\
Flower & $30.10$ & $+0.16$ & $+0.99$ & $-0.48$ & $+0.12$ & $+0.32$ & $+1.49$ & $\mathbf{+1.78}$ \\
Monarch & $28.32$ & $+0.07$ & $+1.32$ & $+1.94$ & $+1.56$ & $+1.39$ & $+2.64$ & $\mathbf{+2.90}$ \\
Leaves & $28.90$ & $+0.26$ & $+1.70$ & $+1.17$ & $+0.70$ & $+1.69$ & $+3.14$ & $\mathbf{+3.44}$ \\
Barbara & $30.41$ & $+0.15$ & $+1.36$ & $+1.43$ & $-1.19$ & $+1.20$ & $+2.81$ & $\mathbf{+3.25}$ \\
Boat & $31.65$ & $+0.34$ & $+1.17$ & $-0.94$ & $-0.27$ & $+0.60$ & $+1.55$ & $\mathbf{+1.84}$ \\
House & $33.72$ & $+0.39$ & $+0.95$ & $-13.98$ & $+0.10$ & $+0.14$ & $\mathbf{+1.70}$ & $+1.60$ \\
Bike & $27.22$ & $+0.08$ & $+1.03$ & $+1.21$ & $+0.06$ & $+0.87$ & $+1.98$ & $\mathbf{+2.27}$ \\
\hline\hline
Median & $30.53$ & $+0.22$ & $+1.01$ & $+0.09$ & $+0.11$ & $+0.54$ & $+1.62$ & $\mathbf{+1.81}$ \\
\hline
\end{tabular}
\caption{PSNR gains (dB) of different denoising algorithms at $\mbox{\gls{qf}}=25$.}
\label{tab:qf25psnr}
\end{table}

\begin{table}
\centering\footnotesize\begin{tabular}{|c|c|c|c|c|c|c|c|c|}
\hline
Image & JPEG & ACR & BM3D & WNNM & TV & DicTV & DTPD & Proposed \\
\hline\hline
Lenna & $32.96$ & $+0.05$ & $+0.80$ & $+0.85$ & $-0.05$ & $+0.06$ & $+1.44$ & $\mathbf{+1.70}$ \\
Parrot & $34.80$ & $+0.10$ & $+0.64$ & $-0.49$ & $-0.39$ & $-0.11$ & $+1.39$ & $\mathbf{+1.63}$ \\
Hat & $33.64$ & $+0.05$ & $+0.89$ & $+0.01$ & $+0.01$ & $-0.01$ & $+1.60$ & $\mathbf{+1.83}$ \\
Flower & $32.43$ & $+0.02$ & $+0.98$ & $+0.46$ & $+0.12$ & $-0.20$ & $+1.68$ & $\mathbf{+2.08}$ \\
Monarch & $30.73$ & $+0.00$ & $+1.32$ & $+1.99$ & $+1.75$ & $+1.33$ & $+2.62$ & $\mathbf{+3.12}$ \\
Leaves & $31.64$ & $+0.04$ & $+1.78$ & $+0.45$ & $+0.64$ & $+1.58$ & $+3.29$ & $\mathbf{+3.86}$ \\
Barbara & $33.56$ & $+0.05$ & $+1.28$ & $+1.68$ & $-1.75$ & $+0.56$ & $+2.51$ & $\mathbf{+2.98}$ \\
Boat & $34.42$ & $+0.09$ & $+1.20$ & $+0.89$ & $-0.64$ & $-0.11$ & $+1.52$ & $\mathbf{+2.04}$ \\
House & $35.79$ & $+0.13$ & $+0.81$ & $-10.87$ & $-0.11$ & $-0.69$ & $+1.48$ & $\mathbf{+1.52}$ \\
Bike & $29.95$ & $+0.01$ & $+1.10$ & $+1.45$ & $-0.15$ & $+0.84$ & $+2.21$ & $\mathbf{+2.64}$ \\
\hline\hline
Median & $33.26$ & $+0.05$ & $+1.04$ & $+0.66$ & $-0.08$ & $+0.03$ & $+1.64$ & $\mathbf{+2.06}$ \\
\hline
\end{tabular}
\caption{PSNR gains (dB) of different denoising algorithms at $\mbox{\gls{qf}}=50$.}
\label{tab:qf50psnr}
\end{table}

\begin{table}
\centering\footnotesize\begin{tabular}{|c|c|c|c|c|c|c|c|c|}
\hline
Image & JPEG & ACR & BM3D & WNNM & TV & DicTV & DTPD & Proposed \\
\hline\hline
Lenna & $36.59$ & $+0.00$ & $+0.57$ & $+0.53$ & $-0.69$ & $-1.25$ & $+1.22$ & $\mathbf{+1.57}$ \\
Parrot & $38.31$ & $+0.01$ & $+0.46$ & $+0.49$ & $-0.99$ & $-1.45$ & $+1.07$ & $\mathbf{+1.47}$ \\
Hat & $37.27$ & $+0.02$ & $+0.77$ & $+0.96$ & $-0.71$ & $-1.41$ & $+1.65$ & $\mathbf{+2.14}$ \\
Flower & $36.20$ & $+0.00$ & $+0.98$ & $+1.29$ & $-0.57$ & $-1.56$ & $+1.71$ & $\mathbf{+2.32}$ \\
Monarch & $34.73$ & $-0.00$ & $+1.22$ & $+1.87$ & $+1.01$ & $+0.11$ & $+2.33$ & $\mathbf{+3.10}$ \\
Leaves & $35.92$ & $-0.01$ & $+1.74$ & $+2.63$ & $-0.30$ & $-0.02$ & $+3.10$ & $\mathbf{+3.97}$ \\
Barbara & $37.61$ & $+0.00$ & $+0.89$ & $+1.20$ & $-2.59$ & $-1.56$ & $+1.69$ & $\mathbf{+2.09}$ \\
Boat & $38.38$ & $+0.01$ & $+1.04$ & $+1.21$ & $-1.55$ & $-2.29$ & $+1.04$ & $\mathbf{+1.77}$ \\
House & $39.11$ & $+0.01$ & $+0.87$ & $+0.90$ & $-0.88$ & $-2.52$ & $+1.70$ & $\mathbf{+1.98}$ \\
Bike & $34.53$ & $+0.00$ & $+1.12$ & $+1.63$ & $-1.19$ & $-0.52$ & $+2.03$ & $\mathbf{+2.70}$ \\
\hline\hline
Median & $36.93$ & $+0.00$ & $+0.94$ & $+1.21$ & $-0.80$ & $-1.43$ & $+1.69$ & $\mathbf{+2.12}$ \\
\hline
\end{tabular}
\caption{PSNR gains (dB) of different denoising algorithms at $\mbox{\gls{qf}}=80$.}
\label{tab:qf80psnr}
\end{table}

\begin{table}
\centering\footnotesize\begin{tabular}{|c|c|c|c|c|c|c|c|c|}
\hline
Image & JPEG & ACR & BM3D & WNNM & TV & DicTV & DTPD & Proposed \\
\hline\hline
Lenna & $0.8835$ & $+0.0097$ & $+0.0231$ & $+0.0177$ & $+0.0041$ & $+0.0050$ & $+0.0260$ & $\mathbf{+0.0310}$ \\
Parrot & $0.9060$ & $+0.0114$ & $+0.0186$ & $-0.0141$ & $+0.0063$ & $+0.0032$ & $+0.0193$ & $\mathbf{+0.0243}$ \\
Hat & $0.8752$ & $+0.0094$ & $+0.0221$ & $+0.0021$ & $+0.0058$ & $+0.0058$ & $+0.0272$ & $\mathbf{+0.0315}$ \\
Flower & $0.8816$ & $+0.0080$ & $+0.0271$ & $+0.0123$ & $+0.0040$ & $+0.0045$ & $+0.0306$ & $\mathbf{+0.0382}$ \\
Monarch & $0.8969$ & $+0.0069$ & $+0.0425$ & $+0.0453$ & $+0.0396$ & $+0.0371$ & $+0.0514$ & $\mathbf{+0.0536}$ \\
Leaves & $0.9234$ & $+0.0112$ & $+0.0386$ & $+0.0431$ & $+0.0302$ & $+0.0362$ & $+0.0472$ & $\mathbf{+0.0487}$ \\
Barbara & $0.9033$ & $+0.0079$ & $+0.0245$ & $+0.0210$ & $-0.0255$ & $+0.0078$ & $+0.0348$ & $\mathbf{+0.0395}$ \\
Boat & $0.8905$ & $+0.0102$ & $+0.0254$ & $+0.0188$ & $-0.0004$ & $+0.0045$ & $+0.0293$ & $\mathbf{+0.0340}$ \\
House & $0.8741$ & $+0.0060$ & $+0.0110$ & $-0.0431$ & $+0.0006$ & $+0.0024$ & $\mathbf{+0.0168}$ & $+0.0166$ \\
Bike & $0.8798$ & $+0.0055$ & $+0.0259$ & $+0.0180$ & $+0.0019$ & $+0.0134$ & $+0.0401$ & $\mathbf{+0.0458}$ \\
\hline\hline
Median & $0.8870$ & $+0.0087$ & $+0.0250$ & $+0.0179$ & $+0.0040$ & $+0.0054$ & $+0.0299$ & $\mathbf{+0.0361}$ \\
\hline
\end{tabular}
\caption{SSIM gains of different denoising algorithms at $\mbox{\gls{qf}}=25$.}
\label{tab:qf25ssim}
\end{table}

\begin{table}
\centering\footnotesize\begin{tabular}{|c|c|c|c|c|c|c|c|c|}
\hline
Image & JPEG & ACR & BM3D & WNNM & TV & DicTV & DTPD & Proposed \\
\hline\hline
Lenna & $0.9221$ & $+0.0021$ & $+0.0107$ & $+0.0071$ & $-0.0011$ & $-0.0094$ & $+0.0144$ & $\mathbf{+0.0186}$ \\
Parrot & $0.9374$ & $+0.0033$ & $+0.0084$ & $-0.0031$ & $-0.0017$ & $-0.0108$ & $+0.0096$ & $\mathbf{+0.0134}$ \\
Hat & $0.9182$ & $+0.0032$ & $+0.0142$ & $+0.0070$ & $-0.0025$ & $-0.0095$ & $+0.0200$ & $\mathbf{+0.0236}$ \\
Flower & $0.9253$ & $+0.0014$ & $+0.0174$ & $+0.0107$ & $+0.0029$ & $-0.0076$ & $+0.0218$ & $\mathbf{+0.0277}$ \\
Monarch & $0.9300$ & $+0.0009$ & $+0.0282$ & $+0.0294$ & $+0.0278$ & $+0.0229$ & $+0.0350$ & $\mathbf{+0.0377}$ \\
Leaves & $0.9533$ & $+0.0013$ & $+0.0248$ & $+0.0240$ & $+0.0203$ & $+0.0211$ & $+0.0299$ & $\mathbf{+0.0322}$ \\
Barbara & $0.9456$ & $+0.0019$ & $+0.0130$ & $+0.0132$ & $-0.0174$ & $-0.0056$ & $+0.0181$ & $\mathbf{+0.0215}$ \\
Boat & $0.9319$ & $+0.0029$ & $+0.0165$ & $+0.0145$ & $-0.0017$ & $-0.0069$ & $+0.0189$ & $\mathbf{+0.0237}$ \\
House & $0.9103$ & $+0.0015$ & $+0.0040$ & $-0.0369$ & $-0.0085$ & $-0.0166$ & $+0.0100$ & $\mathbf{+0.0124}$ \\
Bike & $0.9280$ & $+0.0008$ & $+0.0162$ & $+0.0130$ & $+0.0022$ & $+0.0005$ & $+0.0276$ & $\mathbf{+0.0321}$ \\
\hline\hline
Median & $0.9290$ & $+0.0017$ & $+0.0152$ & $+0.0118$ & $-0.0014$ & $-0.0072$ & $+0.0194$ & $\mathbf{+0.0237}$ \\
\hline
\end{tabular}
\caption{SSIM gains of different denoising algorithms at $\mbox{\gls{qf}}=50$.}
\label{tab:qf50ssim}
\end{table}

\begin{table}
\centering\footnotesize\begin{tabular}{|c|c|c|c|c|c|c|c|c|}
\hline
Image & JPEG & ACR & BM3D & WNNM & TV & DicTV & DTPD & Proposed \\
\hline\hline
Lenna & $0.9559$ & $+0.0001$ & $+0.0015$ & $-0.0018$ & $-0.0045$ & $-0.0198$ & $+0.0061$ & $\mathbf{+0.0079}$ \\
Parrot & $0.9629$ & $+0.0003$ & $+0.0018$ & $-0.0001$ & $-0.0041$ & $-0.0192$ & $+0.0041$ & $\mathbf{+0.0061}$ \\
Hat & $0.9580$ & $+0.0010$ & $+0.0053$ & $+0.0031$ & $-0.0060$ & $-0.0214$ & $+0.0109$ & $\mathbf{+0.0131}$ \\
Flower & $0.9623$ & $+0.0001$ & $+0.0090$ & $+0.0089$ & $-0.0004$ & $-0.0156$ & $+0.0119$ & $\mathbf{+0.0150}$ \\
Monarch & $0.9626$ & $+0.0000$ & $+0.0121$ & $+0.0120$ & $+0.0131$ & $+0.0050$ & $+0.0174$ & $\mathbf{+0.0191}$ \\
Leaves & $0.9789$ & $-0.0003$ & $+0.0114$ & $+0.0128$ & $+0.0081$ & $+0.0046$ & $+0.0131$ & $\mathbf{+0.0147}$ \\
Barbara & $0.9738$ & $+0.0001$ & $+0.0034$ & $+0.0028$ & $-0.0116$ & $-0.0169$ & $+0.0056$ & $\mathbf{+0.0073}$ \\
Boat & $0.9659$ & $+0.0002$ & $+0.0066$ & $+0.0065$ & $-0.0053$ & $-0.0204$ & $+0.0069$ & $\mathbf{+0.0101}$ \\
House & $0.9530$ & $+0.0002$ & $+0.0040$ & $+0.0023$ & $-0.0133$ & $-0.0346$ & $+0.0111$ & $\mathbf{+0.0121}$ \\
Bike & $0.9679$ & $+0.0000$ & $+0.0087$ & $+0.0086$ & $-0.0011$ & $-0.0118$ & $+0.0130$ & $\mathbf{+0.0156}$ \\
\hline\hline
Median & $0.9627$ & $+0.0001$ & $+0.0060$ & $+0.0048$ & $-0.0043$ & $-0.0180$ & $+0.0110$ & $\mathbf{+0.0126}$ \\
\hline
\end{tabular}
\caption{SSIM gains of different denoising algorithms at $\mbox{\gls{qf}}=80$.}
\label{tab:qf80ssim}
\end{table}

We select several widely used images in the literature as test images
(thumbnailed in Figure~\ref{fig:testthumb}).  All images are $256
\times 256$ in size.  Tables~\ref{tab:qf25psnr}, \ref{tab:qf50psnr}
and \ref{tab:qf80psnr} list the PSNR results of the compared
algorithms on the test images compressed using JPEG with \gls{qf} set
to 25, 50 and 80, respectively.  As shown in the tables, the proposed
technique improves over the hard-decoded JPEG by around 2dB in PSNR.
It leads in PSNR gain in almost every test case and has more than
0.2dB advantage over the second best method.  As a reference, we also
list objective fidelity assessment results by more sophisticated image
quality metric SSIM \cite{wang2004image} in Tables~\ref{tab:qf25ssim},
\ref{tab:qf50ssim} and \ref{tab:qf80ssim} for different \gls{qf}
settings.  As shown in the tables, the SSIM results also confirm the
superiority of the proposed algorithm over the tested technologies.

\begin{figure}
  \centering
  \includegraphics[width = 0.8\textwidth]{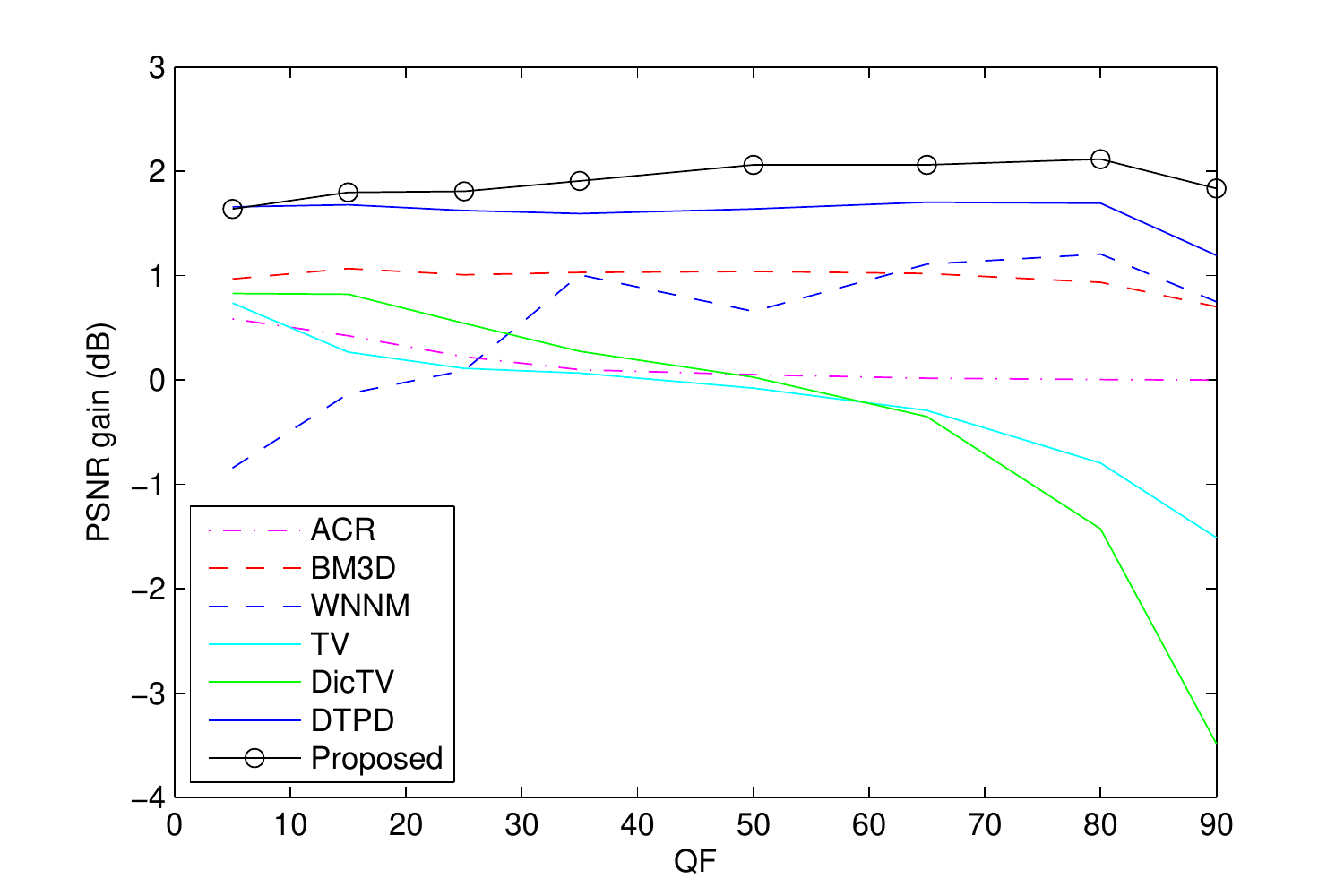}
  \caption{The median PSNR gain as a function of \gls{qf}.}
  \label{fig:psnrqf}
\end{figure}

\begin{figure}
  \centering
  \includegraphics[width = 0.8\textwidth]{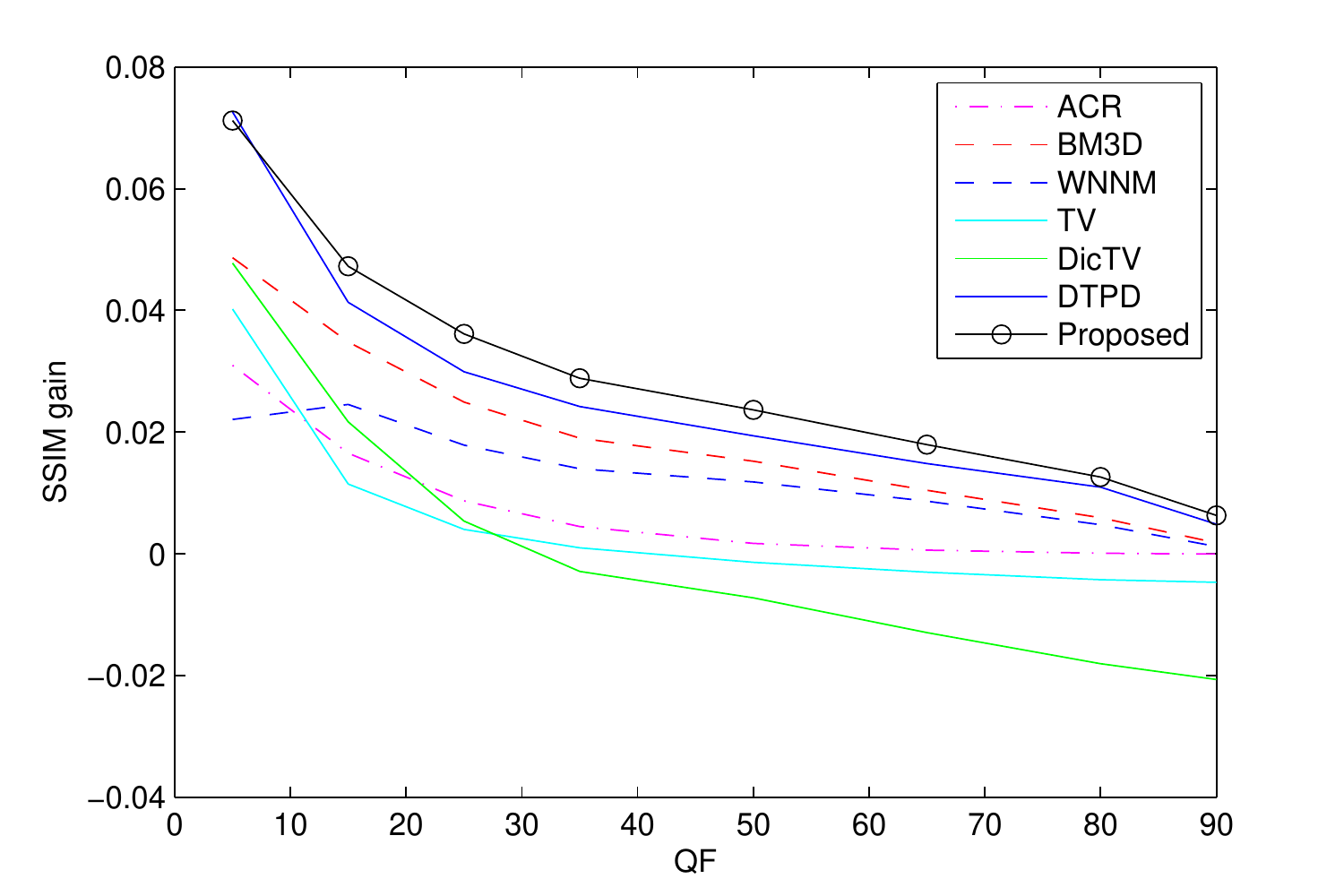}
  \caption{The median SSIM gain as a function of \gls{qf}.}
  \label{fig:ssimqf}
\end{figure}

Compared with other techniques, the proposed technique works
consistently well at vastly different \gls{qf} settings.  As
demonstrated in Figures.~\ref{fig:psnrqf} and \ref{fig:ssimqf}, the
proposed technique is ahead of the competitions at all \gls{qf}
settings except when $\mbox{\gls{qf}}=5$.  Only in that case, the
proposed technique does not perform as well as DTPD in terms of median
PSNR and SSIM gain.  Although $\mbox{\gls{qf}}=5$ is often used in
JPEG denoising research to showcase the capability of a technique, it
has no practical value as compressing a down-scaled version of the
input image with slightly larger \gls{qf} could easily yield better
output image than using $\mbox{\gls{qf}}=5$ directly.  Furthermore, if
we trade off time by increasing the number of iterations $K$, the
proposed technique can outperform DTPD in both PSNR and SSIM while
still being faster than DTPD at $\mbox{\gls{qf}}=5$.

\begin{figure}
  \centering
  \begin{tabular}{c@{ }c@{ }c@{ }c}
    \small JPEG & \small ACR & \small BM3D & \small WNNM \\
    \includegraphics[width=0.24 \textwidth]{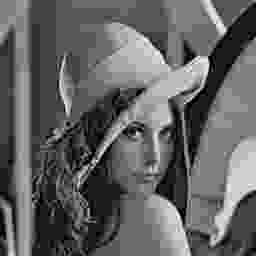}
    & \includegraphics[width=0.24 \textwidth]{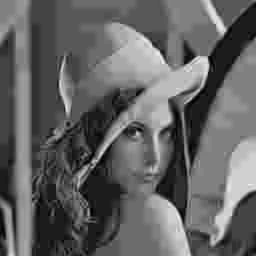}
    & \includegraphics[width=0.24 \textwidth]{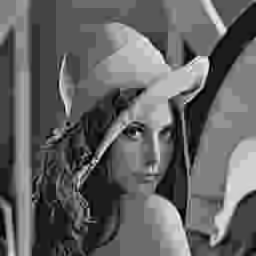}
    & \includegraphics[width=0.24 \textwidth]{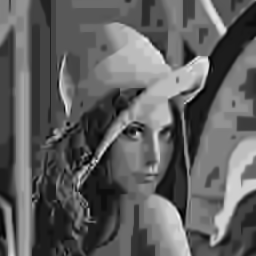} \\
    \small TV & \small DicTV & \small DTPD & \small Proposed \\
    \includegraphics[width=0.24 \textwidth]{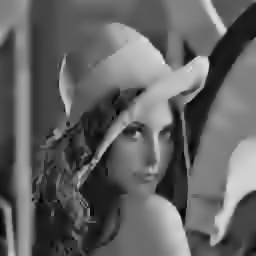}
    & \includegraphics[width=0.24 \textwidth]{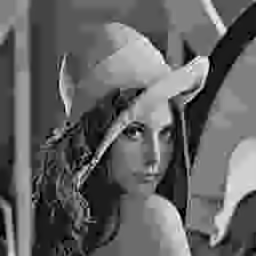}
    & \includegraphics[width=0.24 \textwidth]{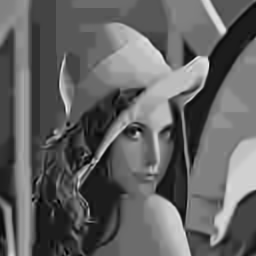}
    & \includegraphics[width=0.24 \textwidth]{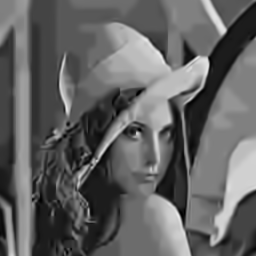} \\
  \end{tabular}
  \caption{Comparison of tested methods in visual quality at
    $\mbox{\gls{qf}}=5$.}
  \label{fig:subqf5}
\end{figure}

\begin{figure}
  \centering
  \begin{tabular}{c@{ }c@{ }c@{ }c}
    \small JPEG & \small ACR & \small BM3D & \small WNNM \\
    \includegraphics[width=0.24 \textwidth]{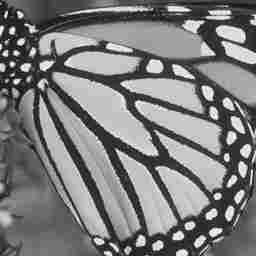}
    & \includegraphics[width=0.24 \textwidth]{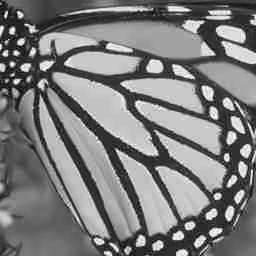}
    & \includegraphics[width=0.24 \textwidth]{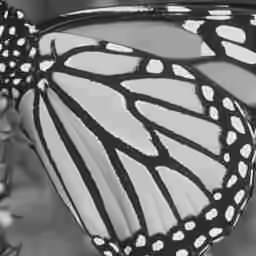}
    & \includegraphics[width=0.24 \textwidth]{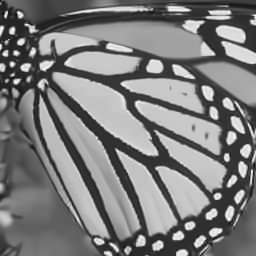} \\
    \small TV & \small DicTV & \small DTPD & \small Proposed \\
    \includegraphics[width=0.24 \textwidth]{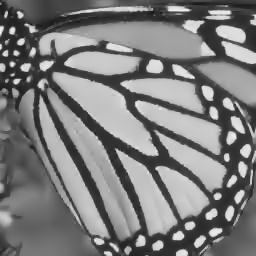}
    & \includegraphics[width=0.24 \textwidth]{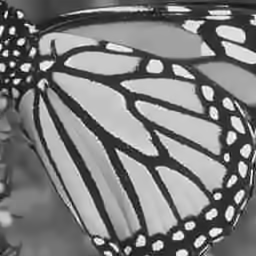}
    & \includegraphics[width=0.24 \textwidth]{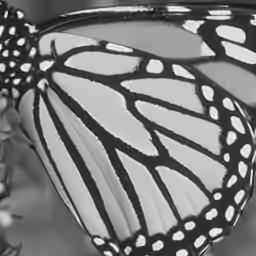}
    & \includegraphics[width=0.24 \textwidth]{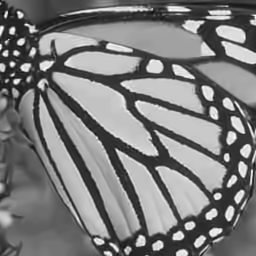} \\
  \end{tabular}
  \caption{Comparison of tested methods in visual quality at
    $\mbox{\gls{qf}}=15$.}
  \label{fig:subqf15}
\end{figure}

\begin{figure}
  \centering
  \begin{tabular}{c@{ }c@{ }c@{ }c}
    \small JPEG & \small ACR & \small BM3D & \small WNNM \\
    \includegraphics[width=0.24 \textwidth]{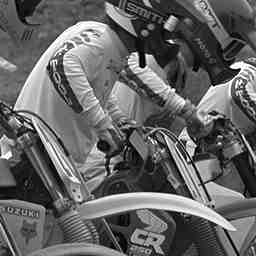}
    & \includegraphics[width=0.24 \textwidth]{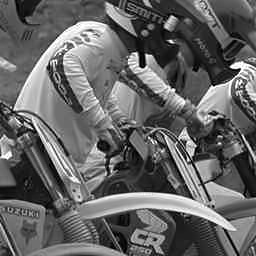}
    & \includegraphics[width=0.24 \textwidth]{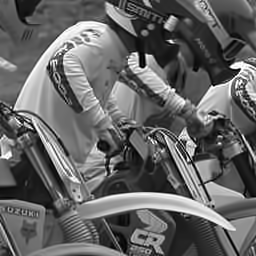}
    & \includegraphics[width=0.24 \textwidth]{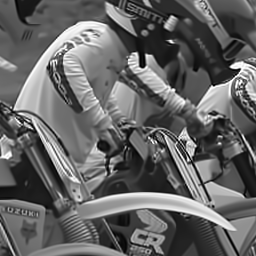} \\
    \small TV & \small DicTV & \small DTPD & \small Proposed \\
    \includegraphics[width=0.24 \textwidth]{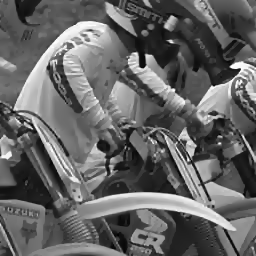}
    & \includegraphics[width=0.24 \textwidth]{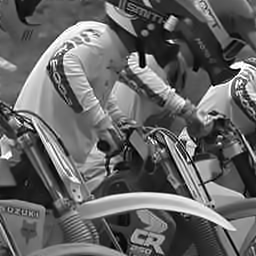}
    & \includegraphics[width=0.24 \textwidth]{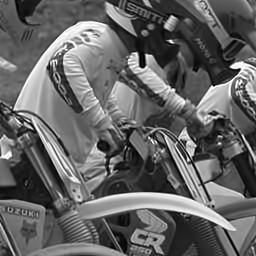}
    & \includegraphics[width=0.24 \textwidth]{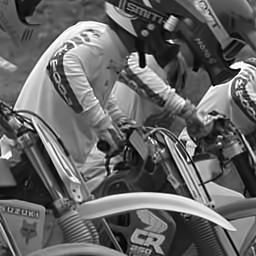} \\
  \end{tabular}
  \caption{Comparison of tested methods in visual quality at
    $\mbox{\gls{qf}}=25$.}
  \label{fig:subqf25}
\end{figure}


In addition to its superior performance in objective fidelity metric,
the proposed approach also obtains better perceptual quality of the
denoised images.  As shown in Figures~\ref{fig:subqf5},
\ref{fig:subqf15} and \ref{fig:subqf25} are some samples of the
results from the tested algorithms.  The output images of the proposed
approach shows no discernible blocking and ringing artifacts even at
low QF settings.  The proposed approach preserves detail and edge
structure visibly better than most of other techniques.

Like all the tested techniques except BM3D whose main functions are
implemented in C++ and compiled to native code, the reference
implementation of the proposed technique is written in pure MATLAB
language, rendering it unfavourable in comparison of time cost with
BM3D.  Besides BM3D, the only other method faster than the proposed
technique in the comparison group is ACR, which only reduces blocking
artifacts and does not perform as well as most of the compared
techniques in terms of either PSNR or SSIM.

\section{Conclusion}

Due to the low pass nature of image compression, the high-frequency
components of a compressed image with sharp edges often carry large
compression error.  While high-frequency compression noise is
relatively indiscernible in the original image as \gls{hvs} is more
sensitive to low-frequency noise, image restoration operator with
high-boosting property can amplify the problem deteriorating the
perceptive quality of restored image.  By incorporating the non-linear
DCT quantization mechanism into the formulation for image restoration,
we propose new sparsity-based convex programming approach for joint
quantization noise removal and restoration.  Experimental results
demonstrate significant performance gains of the new approach over
existing restoration methods.

{\small
\bibliographystyle{ieee}
\bibliography{imgrest}
}

\end{document}